\documentclass{article}
\PassOptionsToPackage{numbers, compress}{natbib}
\usepackage[preprint]{neurips_2026}

\usepackage[utf8]{inputenc} 
\usepackage[T1]{fontenc}    
\usepackage{hyperref}       
\usepackage{url}            
\usepackage{booktabs}       
\usepackage{amsfonts}       
\usepackage{amsmath}
\usepackage{cleveref}
\usepackage{nicefrac}       
\usepackage{microtype}      
\usepackage{xcolor}         
\usepackage{algorithm}
\usepackage{algorithmic}
\usepackage{arxiv}
\usepackage{tabularx}
\usepackage{array}
\usepackage{xcolor}
\usepackage{silence}
\title{TOC-Bench: A Temporal Object Consistency Benchmark for Video Large Language Models}
\author{Junzhe Chen\thanks{\texttt{junzhec@tju.edu.cn}}, Siyuan Meng, Yuxi Chen, Man Zhao, Wenyao Gui, and Xiaojie Guo\thanks{\texttt{xj.max.guo@gmail.com}}\\
Tianjin University
}

\begin{document}

\maketitle

\begin{abstract}
Video large language models (Video-LLMs) have achieved remarkable progress in general video understanding, yet their ability to maintain \emph{temporal object consistency} remains insufficiently explored. Existing benchmarks primarily focus on event recognition, action understanding, or coarse temporal reasoning, but rarely evaluate whether a model can consistently preserve the identity, state, and temporal continuity of the same object across occlusion, disappearance, reappearance, state transitions, and cross-object interactions. As a result, current evaluations may overestimate temporal reasoning ability while overlooking failures in object-centric temporal coherence.
To address this issue, we introduce \emph{TOC-Bench}, a diagnostic benchmark specifically designed to evaluate temporal object consistency in Video-LLMs. TOC-Bench is explicitly \emph{object-track grounded}, where each queried subject is associated with a per-frame object trajectory and structured temporal event timeline. To ensure that benchmark items depend on temporally ordered visual evidence rather than language priors, single-frame shortcuts, or unordered frame cues, we propose a three-layer \emph{temporal-necessity filtering} protocol that removes 60.7\% of candidate QA pairs and retains 17,900 temporally dependent items spanning 10 diagnostic dimensions. From this filtered pool, we further construct a human-verified benchmark containing 2,323 high-quality QA pairs over 1,951 videos.
Experiments on representative Video-LLMs show that temporal object consistency remains a major unsolved challenge. Current models exhibit substantial weaknesses in event counting, event ordering, identity-sensitive reasoning, and hallucination-aware verification, despite strong performance on general video understanding benchmarks. These findings suggest that temporal object consistency is a major bottleneck for current Video-LLMs, and TOC-Bench offers a focused platform for diagnosing and improving object-aware temporal reasoning. The resource is available at \url{https://github.com/cjzcjz666/toc_bench.git}.

\end{abstract}

\section{Introduction}

Video-LLMs have made rapid progress in video understanding, enabling models to answer questions, describe events, localize moments, and reason over long and complex video inputs. Early video-instruction models such as VideoChatGPT, Video-LLaVA, VideoChat, Video-LLaMA, and LLaVA-NeXT-Video established effective paradigms for video-language instruction tuning~\cite{maaz2024video, lin2024video, li2025videochat, cheng2024videollama, zhang2024llavanext-video}. More recent multimodal systems further improve long-context video perception and general multimodal reasoning~\cite{wang2024qwen2, chen2024internvl, wang2025internvideo2, ryoo2024xgen, Qwen2.5-VL}. Meanwhile, video understanding benchmarks have expanded from short-video QA and activity understanding~\cite{xu2016msr, xu2017video, yu2019activitynet} to broad temporal reasoning, long-video comprehension, and shortcut-aware evaluation~\cite{li2024mvbench, fu2025video, fang2024mmbench, wu2024longvideobench, cores2024tvbench, fu2026video}. These advances raise a finer-grained evaluation question: can Video-LLMs maintain object-level consistency?

We refer to this capability as \emph{temporal object consistency}: the ability to maintain the identity, state, and persistence of the same object throughout a video. This capability is important for real-world video understanding, where objects may be hidden, leave the scene, reappear, or interact over time. For instance, distinguishing whether an object disappears because it is occluded or leaves the frame, identifying whether a reappearing object is the same instance, and determining whether one object's state change happens before another object's event all require temporally ordered evidence anchored to specific objects. Without this capability, a model may correctly recognize individual events but still fail to reason about object persistence and object-level temporal relations.

Despite its importance, temporal object consistency remains insufficiently evaluated as a dedicated evaluation axis. Existing benchmarks have substantially advanced broad temporal understanding, long-form reasoning, and comprehensive video evaluation, but they are often built from video-level annotations, captions, or human-written questions rather than explicit object tracks and event timelines. Recent shortcut-aware benchmarks further show that many video-language questions can be solved from static frames or question-answer text alone~\cite{cores2024tvbench, liu2024tempcompass, krojer2025shortcut}, but they mostly operate at the video-text, event, or temporal-concept level. This leaves an gap in Video-LLM evaluation: existing benchmarks do not directly test whether models can preserve object identity, state, and temporal relations through occlusion, disappearance, reappearance, and cross-object interactions.

\begin{figure}[t]
    \centering
    \includegraphics[width=1\linewidth]{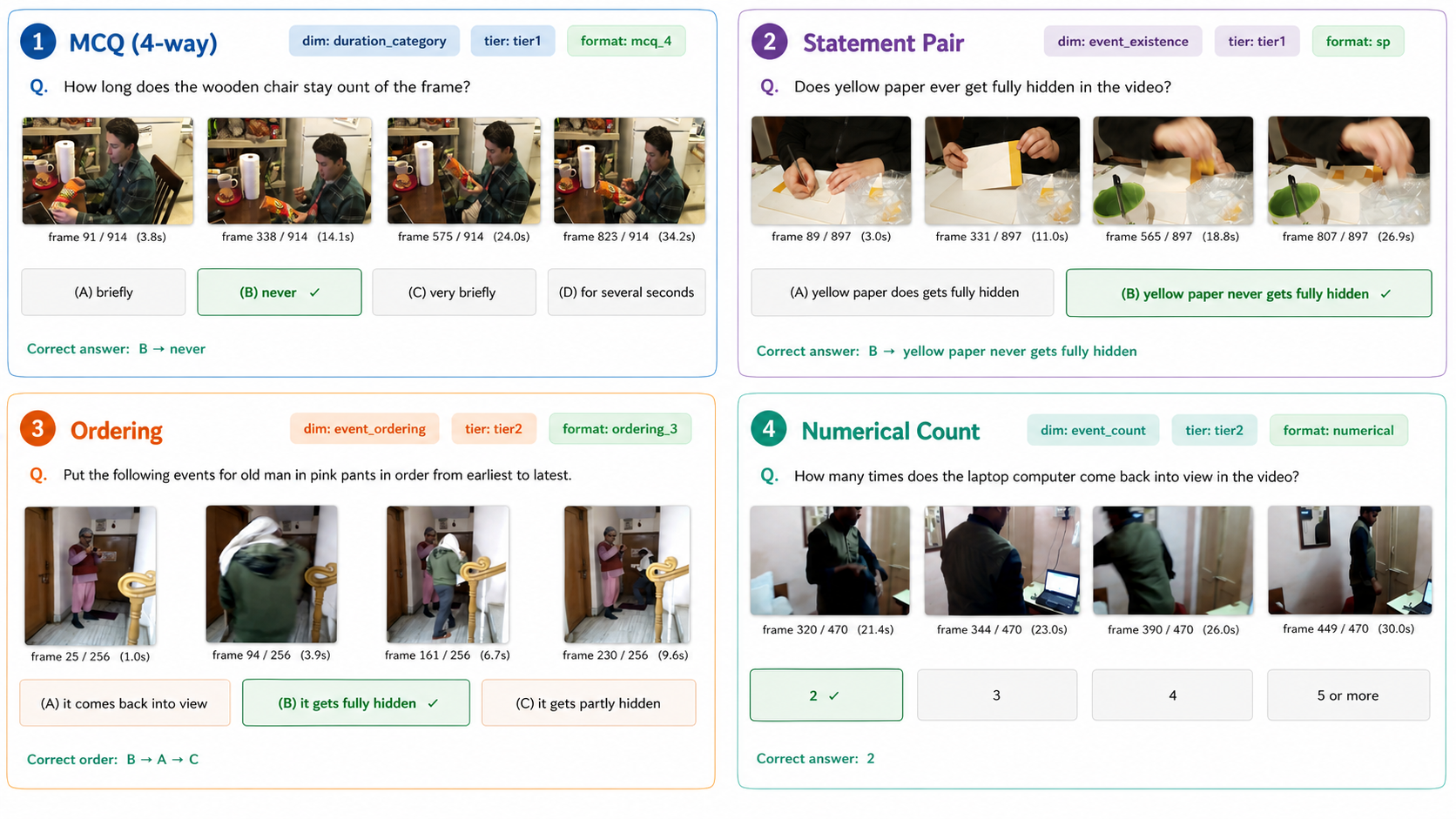}
    \vspace{-15pt}
    \caption{
    Representative TOC-Bench QA examples.
    The benchmark supports multiple deterministic task formats, including four-way multiple choice, statement-pair judgment, event ordering, and numerical counting.
    Correct answers are highlighted in green.
    }
    \vspace{-15pt}
    \label{fig:intro_vis}
\end{figure}

To address this gap, we introduce TOC-Bench, a diagnostic benchmark for evaluating temporal object consistency in Video-LLMs. TOC-Bench is \emph{object-track grounded}: each queried subject is anchored to an explicit per-frame object track, and each QA item is derived from structured object-event timelines rather than free-form object mentions. It contains 2,323 human-verified QA pairs over 1,951 real videos, covers 10 diagnostic dimensions in three difficulty tiers, and supports four deterministic task formats: four-way multiple choice, statement pair, numerical count, and event ordering. TOC-Bench further applies temporal-necessity filtering and includes hallucination-aware variants to reduce shortcut-solvable questions and test whether models over-assume nonexistent objects or events, complementing recent efforts to evaluate video-language hallucination~\cite{li2025vidhalluc, wang2024videohallucer}. Representative examples across the supported task formats are shown in \cref{fig:intro_vis}.

We evaluate 23 representative Video-LLMs on TOC-Bench and find that temporal object consistency remains a significant bottleneck. Current models remain far below human performance, especially on event counting, event ordering, identity-sensitive reasoning, and hallucination-aware verification. These findings suggest that strong general video understanding models do not yet imply reliable object-level temporal reasoning.

Our contributions are summarized as follows:

\begin{itemize}
    \item We introduce \textbf{temporal object consistency} as a dedicated evaluation capability for Video-LLMs, distinct from event-level or video-level temporal reasoning. TOC-Bench evaluates this capability through 10 diagnostic dimensions in three difficulty tiers.

    \item We instantiate this axis as \textbf{TOC-Bench}, an object-track grounded diagnostic benchmark in which each queried subject is tied to a per-frame object track and each answer is derived from a structured object-event timeline. The benchmark contains 2,323 human-verified QA pairs over 1,951 videos and supports four deterministic task formats.


    \item We propose a \textbf{shortcut-aware and hallucination-aware diagnostic protocol}. Starting from 45,527 generated QA pairs, the temporal-necessity filter removes items answerable from text-only, single-frame, or shuffled-frame settings and retains 17,900 temporally dependent candidates. We further integrate hallucination-aware variants and report hallucination diagnostic accuracy (HDA) to evaluate object/event hallucination robustness.
\end{itemize}

\section{Related Work}
\label{sec:related-work}

\subsection{Video Large Language Models}

Recent Video-LLMs extend VLMs to video understanding through video-specific instruction tuning, long-context modeling, and large-scale multimodal foundation models. Representative models such as VideoChatGPT~\cite{maaz2024video}, Video-LLaVA~\cite{lin2024video}, VideoChat~\cite{li2025videochat}, VideoLLaMA~2~\cite{cheng2024videollama}, LLaVA-NeXT-Video~\cite{zhang2024llavanext-video}, and LLaVA-Video~\cite{zhang2024llava} adapt image-language models to video QA, captioning, and dialogue. More recent models, including VideoChat-Flash~\cite{li2024videochat}, InternVideo2/2.5~\cite{wang2024internvideo2, wang2025internvideo2}, Qwen2.5-VL~\cite{Qwen2.5-VL}, Qwen3-VL~\cite{bai2025qwen3}, InternVL3~\cite{zhu2025internvl3}, and Tarsier2~\cite{yuan2025tarsier2}, further improve long-context perception and general multimodal reasoning. Despite these advances, many Video-LLMs still rely on frame sampling, visual-token compression, or global temporal aggregation, which can weaken the object-level continuity needed for temporal object consistency.

\subsection{Video QA Benchmarks and Temporal Evaluation}

Video QA benchmarks have evolved from short-video QA datasets such as MSVD-QA~\cite{xu2017video}, MSRVTT-QA~\cite{xu2016msr}, ActivityNet-QA~\cite{yu2019activitynet}, TGIF-QA~\cite{jang2017tgif}, NExT-QA~\cite{xiao2021next}, and STAR~\cite{wu2024star}, to broader Video-LLM evaluation suites such as EgoSchema~\cite{mangalam2023egoschema}, MVBench~\cite{li2024mvbench}, Video-MME~\cite{fu2025video}, Video-MME-v2~\cite{fu2026video}, MMBench-Video~\cite{fang2024mmbench}, LongVideoBench~\cite{wu2024longvideobench}, and LVBench~\cite{wang2025lvbench}. These benchmarks cover activity understanding, long-form comprehension, multi-domain QA, and broad temporal reasoning, but most are built from video-level annotations, captions, or human-written questions rather than explicit object identity tracks.

Recent benchmarks have also examined whether Video-LLMs rely on shortcuts when answering temporal questions. TempCompass~\cite{liu2024tempcompass}, TVBench~\cite{cores2024tvbench}, TemporalBench~\cite{cai2024temporalbench}, and RTV-Bench~\cite{xun2025rtv} reduce reliance on static frames, text priors, or weak temporal supervision. Among them, VITATECS~\cite{li2024vitatecs} is the closest to our work in its diagnostic motivation: it evaluates temporal concept understanding using counterfactual video-text descriptions. TOC-Bench differs by shifting the diagnostic unit from video-level temporal concepts to tracked object instances. Each queried subject in TOC-Bench is grounded in a per-frame object track, and each question is derived from object-specific event timelines rather than from video-text description pairs.

\subsection{Object-Centric Video Grounding and Tracking}

Object-centric video grounding has been advanced by segmentation, tracking, and referring video object segmentation methods such as Segment Anything~\cite{kirillov2023segment}, SAM2~\cite{ravi2024sam2}, Track Anything~\cite{yang2023track}, SAM2Long~\cite{ding2025sam2long}, and VideoRefer Suite~\cite{yuan2025videorefer}. These methods provide tools for localizing, segmenting, and following objects in videos. TOC-Bench uses them as infrastructure rather than as evaluation targets: the benchmark asks whether Video-LLMs can use object-track evidence to answer temporally grounded questions about the same object over time.

\section{TOC-Bench: Temporal Object Consistency Benchmark}
\label{sec:toc-bench}


TOC-Bench evaluates temporal object consistency in Video-LLMs, focusing on whether models can track object identity, state, and persistence across time. We use \emph{diagnostic} to mean that the benchmark isolates this capability through controlled, automatically gradable QA items rather than broad average-case video QA.

TOC-Bench contains 2,323 human-verified QA pairs from 1,951 unique videos across four public sources: Charades~\cite{sigurdsson2016hollywood}, Perception Test~\cite{patraucean2023perception}, OVIS~\cite{qi2022occluded}, and MOSE~\cite{MOSE}. The total video duration is 15 h 37 min, with an average length of 28.8 s. Questions are organized around object grounding (O), temporal evidence (T), and consistency constraints (C), forming three difficulty tiers: Tier 1 (O+T) for single-object temporal reasoning, Tier 2 (O+T+C) for cross-time consistency of one object, and Tier 3 (O+T+C$\times$2) for consistency reasoning across two objects. Overall, TOC-Bench covers 10 diagnostic dimensions and four task formats, including multiple choice, statement pair, numerical count, and event ordering. These dimensions are not meant to cover all forms of video reasoning. Instead, they are designed to capture the core aspects of temporal object consistency, including object persistence, event timing, state change, repeated-event accumulation, identity preservation, relative spatial change, and cross-object temporal relations.

\subsection{Benchmark Construction}
\label{sec:benchmark-construction}

\begin{figure}[t]
    \centering
    \includegraphics[width=1\linewidth]{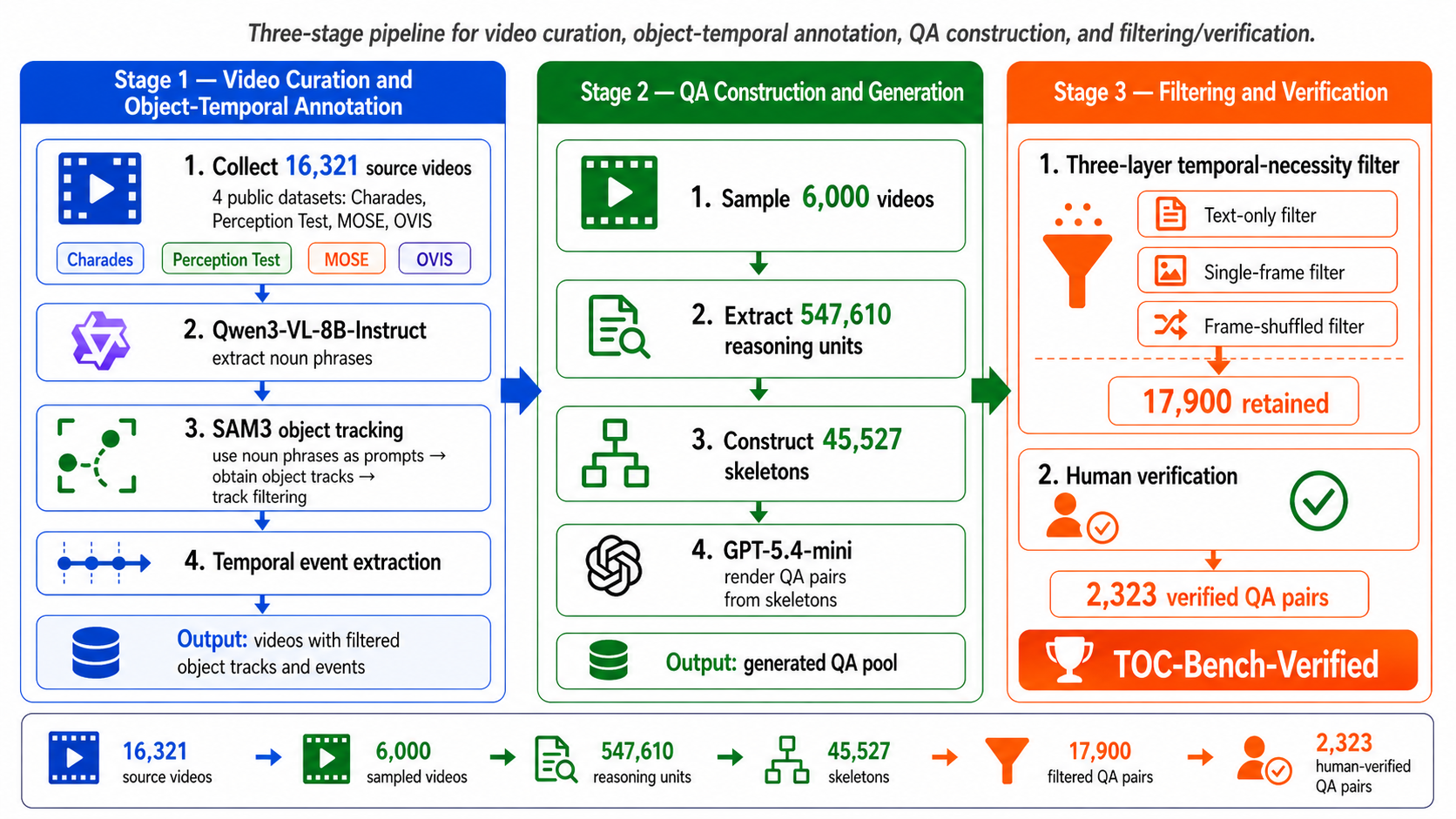}
    \vspace{-10px}
    \caption{The construction pipeline of TOC-Bench.}
    \vspace{-10px}
    \label{fig:pipeline}
\end{figure}

TOC-Bench is built through a three-stage pipeline, as shown in \cref{fig:pipeline}. Stage 1 derives object tracks and event timelines from source videos. Stage 2 converts these structured records into reasoning units and candidate QA pairs. Stage 3 applies temporal-necessity filtering and human verification to obtain the final benchmark.

\begin{figure}[t]
    \centering
    \includegraphics[width=0.6\linewidth]{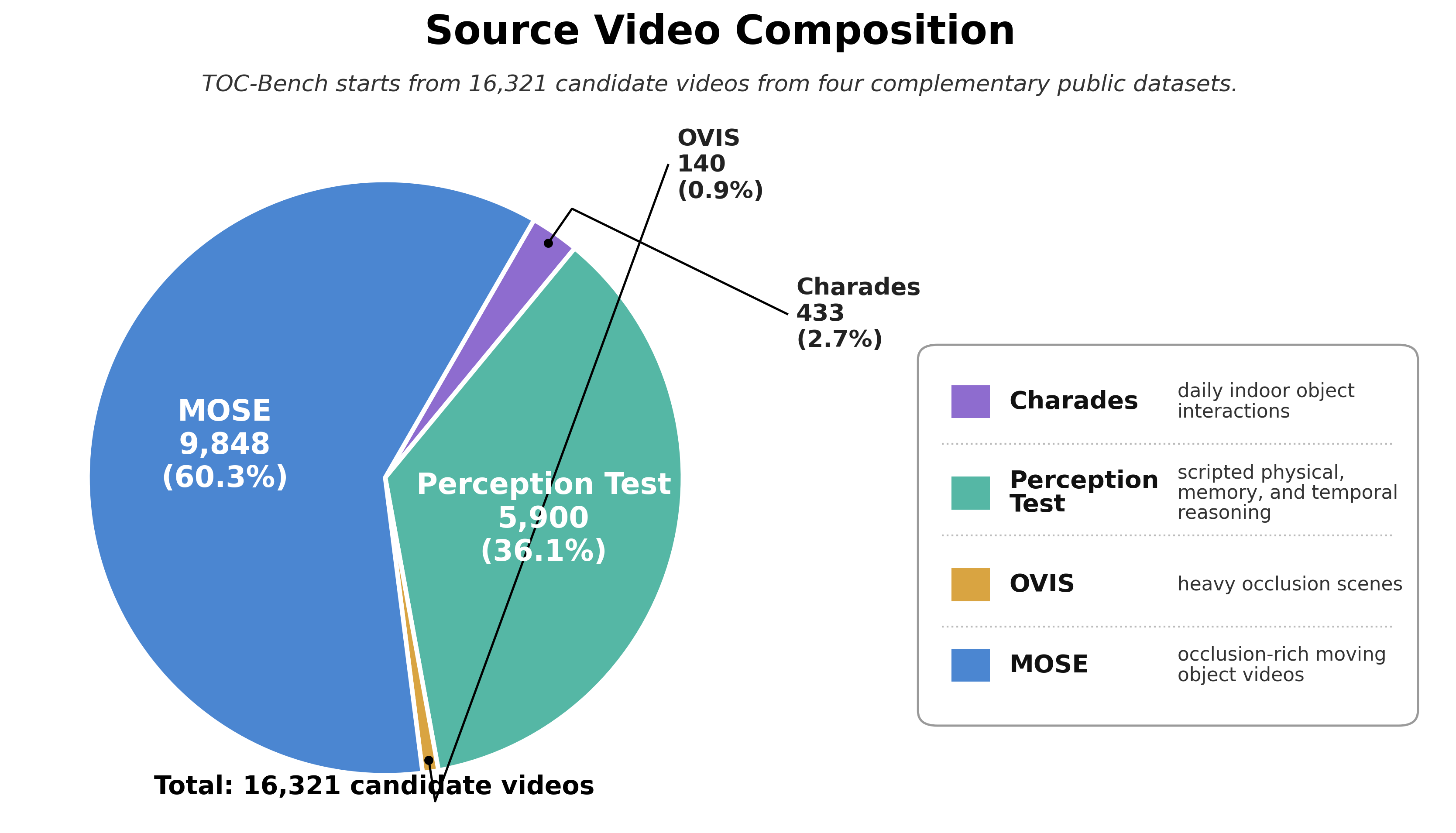}
    \vspace{-5px}
    \caption{Source video composition of TOC-Bench.}
    \vspace{-10px}
    \label{fig:source-video-composition}
\end{figure}

\paragraph{Video curation and object-track annotation.}
As shown in \cref{fig:source-video-composition}, TOC-Bench starts from 16,321 candidate videos from four public datasets. Charades provides daily indoor activities with rich object interactions, while Perception Test contains scripted videos designed for physical, memory, and temporal reasoning. OVIS and MOSE contribute occlusion-heavy videos, which are useful for constructing reappearance, occlusion, and cross-object reasoning questions. Thus, although the QA items are newly constructed, the underlying object-level temporal phenomena occur in real videos rather than synthetic scenes.

For each video, we use Qwen3-VL-8B-Instruct~\cite{bai2025qwen3} with up to 10 evenly sampled frames to propose distinct physical objects and people. The extracted noun phrases are used as prompts for SAM3~\cite{carion2025sam} to obtain candidate per-frame object tracks. We remove low-quality, fragmented, or visually ambiguous tracks, and derive temporal events such as appearance, disappearance, occlusion, reappearance, and cross-object relations from the retained tracks. Qwen3-VL-8B-Instruct and SAM3 are therefore used for scalable candidate annotation, while downstream QA construction relies on filtered object tracks and rule-derived event timelines. More details are provided in \cref{appendix:event-detection}.

\paragraph{QA construction and generation.}
The QA pairs are newly constructed from object tracks and event timelines rather than inherited from the source datasets. From the videos with extracted tracks and events, we sample 6,000 videos for QA construction, prioritizing valid object tracks, clear temporal events, and sufficient object-level changes. For each sampled video, we convert object tracks and event annotations into structured reasoning units. Each reasoning unit specifies the queried object, relevant event or state, diagnostic dimension, answer type, correct answer, and admissible distractors. This step produces 547,610 reasoning units.

We then convert reasoning units into question skeletons. A skeleton fixes the question structure, answer options, correct label, and task format before natural-language generation. From the reasoning units, we construct 45,527 skeletons covering 10 diagnostic dimensions and four task types. GPT-5.4-mini is used only for surface realization: it improves the wording of a pre-specified QA record, while object references, answer labels, temporal buckets, numerical answers, and ordering relations remain fixed by the skeleton. We also add hallucination-aware variants and distractors, including nonexistent subjects, nonexistent events, and plausible but visually unsupported options. Implementation details are provided in \cref{appendix:qa-construction}.

\paragraph{Filtering and verification.}
After QA generation, we apply a three-layer temporal-necessity filter. The text-only filter removes items answerable from the question and options alone. The single-frame filter removes items answerable from individual frames. The frame-shuffled filter removes items whose answers do not depend on the original temporal order. These filters reduce the likelihood that retained questions can be solved from language priors, static visual cues, or unordered frame evidence. The filtering process retains 17,900 of 45,527 generated QA pairs as a candidate pool.


We further sample 3,000 filtered items for human verification. Annotators inspect the corresponding videos, questions, answer candidates, and ground-truth labels. Minor issues such as unclear wording, duplicated options, or slight reference ambiguity are manually revised when the intended evidence and answer remain valid. Items are removed only when they have severe evidence or label problems, such as tracking errors, insufficient visual support, incorrect answers, or a clear mismatch between the QA item and the video. This process yields the final 2,323 human-verified QA pairs.

Upstream models are used only to improve scalable candidate construction and surface realization; final answers are fixed by structured object-event records and retained after rule-based checks and human verification. Additional details on filtering and model-dependence control are provided in \cref{appendix:filtering,appendix:model-dependence}.

\subsection{Dataset Statistics}
\label{sec:dataset-statistics}


TOC-Bench prioritizes diagnostic validity and human verification over raw scale. From 45,527 generated QA pairs, the three-layer filter retains 17,900 temporally dependent candidates, from which we manually verify 2,323 QA pairs over 1,951 videos. Each video contributes 1.19 QA pairs on average, preventing a small number of videos from dominating the benchmark. Charades and Perception Test contribute most items, while MOSE and OVIS provide additional occlusion-heavy cases.

\begin{figure}[t]
    \centering
    \includegraphics[width=1\linewidth]{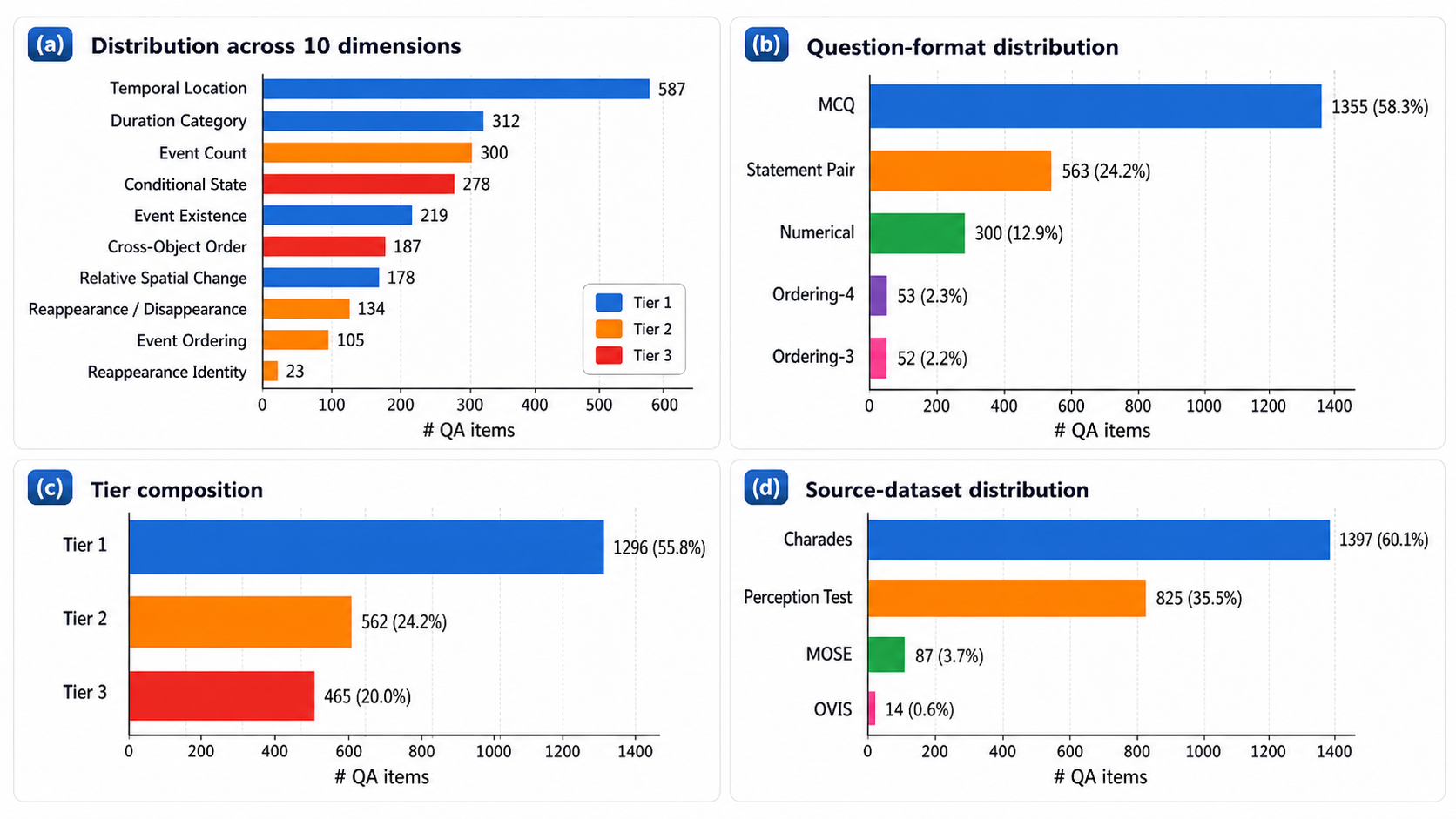}
    \vspace{-5px}
    \caption{
    Overall composition of TOC-Bench.
    (a)-(d) depict the distribution of QA items across the 10 diagnostic dimensions, with colors indicating tier assignments; the question-format distribution; the tier-level composition; and the source-dataset distribution, respectively.
    }
    \vspace{-5px}
    \label{fig:dataset-composition}
\end{figure}

As shown in \cref{fig:dataset-composition}, TOC-Bench covers 10 diagnostic dimensions, with temporal location, duration category, event count, and conditional state forming the largest groups. TOC-Bench intentionally uses deterministic formats rather than open-ended generation, because the goal is reproducible diagnosis of object-level temporal consistency rather than free-form response evaluation. It includes 1,355 multiple-choice questions, 563 statement-pair questions, 300 numerical questions, and 105 event-ordering questions. At the tier level, TOC-Bench contains 1,296 Tier-1 items, 562 Tier-2 items, and 465 Tier-3 items, so 44.2\% of the dataset requires either cross-time consistency for the same object or consistency reasoning across two objects.

\begin{figure}[t]
    \centering
    \includegraphics[width=10cm, height=4.3cm]{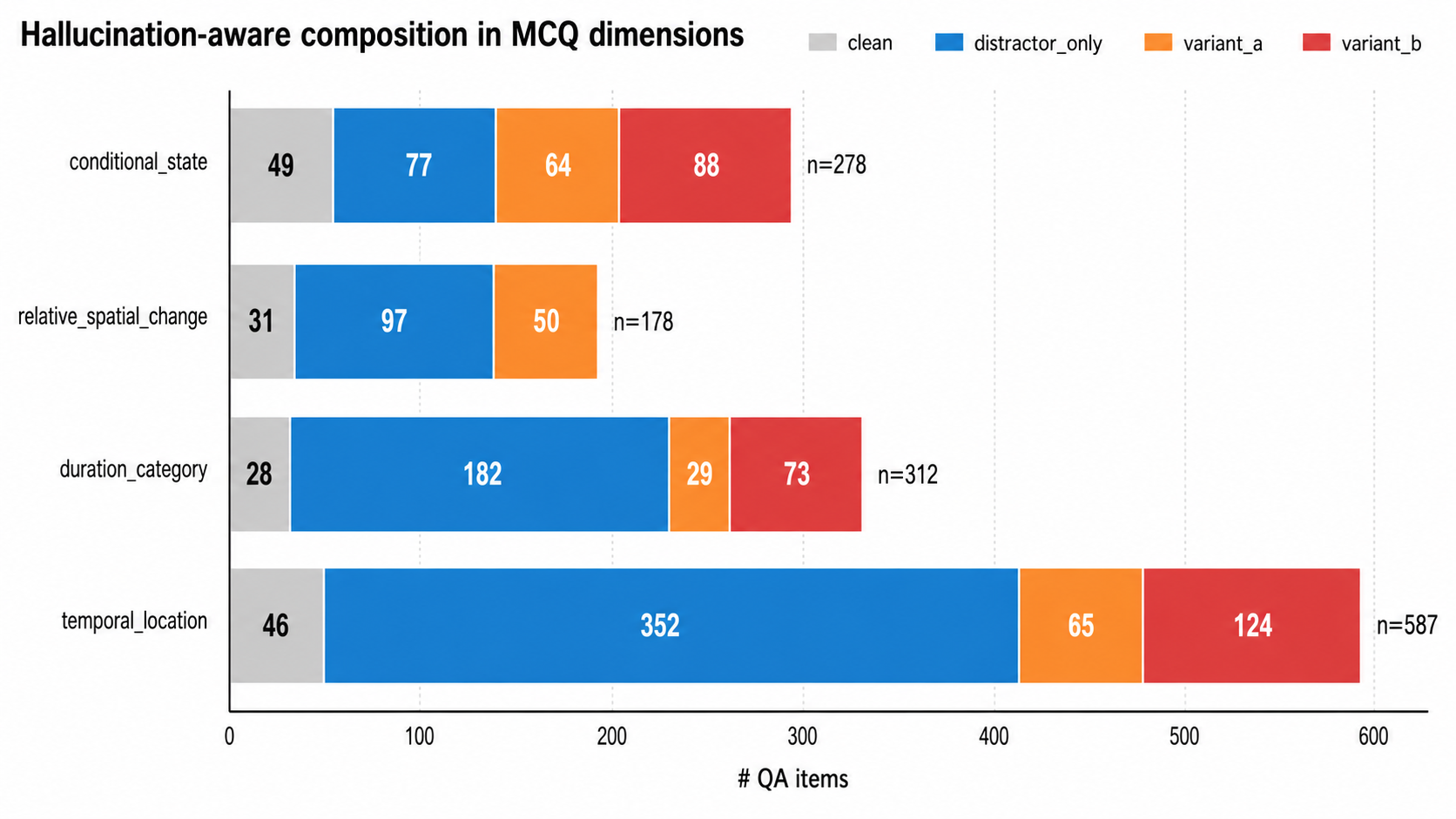}
    \caption{
    Hallucination-aware composition of TOC-Bench in multiple-choice dimensions.
    The stacked bars show the number of clean items, distractor-only items, nonexistent-subject variants (variant A), and nonexistent-event variants (variant B) in each dimension.
    }
    \vspace{-10pt}
    \label{fig:hallucination-composition}
\end{figure}

TOC-Bench includes hallucination-aware items to test whether models can reject plausible but visually unsupported objects or events. As shown in \cref{fig:hallucination-composition}, these items are concentrated in four multiple-choice dimensions: temporal location, duration category, relative spatial change, and conditional state. In addition to clean questions, the benchmark includes distractor-only items, nonexistent-subject variants (Variant A), and nonexistent-event variants (Variant B). This design tests whether a model can avoid inventing objects or events that are not supported by the video.

\begin{table*}[t]
\caption{
Comparison between TOC-Bench and representative video understanding benchmarks.
TOC-Bench is designed as a diagnostic benchmark for temporal object consistency.
``Temporal filter'' indicates whether the benchmark construction or evaluation explicitly controls text-only, single-frame, or frame-order shortcuts.
}
\label{tab:benchmark-comparison}
\centering
\footnotesize
\setlength{\tabcolsep}{5pt}
\renewcommand{\arraystretch}{0.98}
\begin{tabular}{@{}lccp{0.18\textwidth}cccc@{}}
\toprule
Benchmark 
& Videos 
& QA pairs 
& Format 
& \makecell{Object-track\\grounded}
& \makecell{Temporal\\filter}
& \makecell{Halluc.-\\aware}
& \makecell{Auto.\\grading} \\
\midrule
MVBench 
& 4,000 
& 4,000 
& MCQ 
& No 
& No 
& No 
& Yes \\

TVBench 
& -- 
& 2,654 
& MCQ 
& No 
& Yes 
& No 
& Yes \\

EgoSchema 
& 5,000+ 
& 5,000+ 
& MCQ 
& No 
& Partial 
& No 
& Yes \\

VITATECS 
& -- 
& -- 
& Text-video matching 
& No 
& Yes 
& No 
& Yes \\

Video-MME 
& 900 
& 2,700 
& MCQ 
& No 
& No 
& No 
& Yes \\

Video-MME-v2 
& 800 
& 3,200 
& \makecell[l]{MCQ(8-option)}
& No 
& Partial 
& Partial 
& Yes \\

LongVideoBench 
& 3,763 
& 6,678 
& MCQ 
& No 
& Partial 
& No 
& Yes \\

TOC-Bench 
& 1,951 
& 2,323 
& \makecell[l]{MCQ/SP/Num/Ord}
& Yes 
& Yes 
& Yes 
& Yes \\
\bottomrule
\end{tabular}
\end{table*}

As shown in \cref{tab:benchmark-comparison}, TOC-Bench is complementary to existing video understanding benchmarks. Existing benchmarks typically emphasize broad temporal reasoning, long-form understanding, temporal concept discrimination, or multi-domain video QA. TOC-Bench instead focuses on temporal object consistency and differs by combining object-track grounded subjects, temporal-necessity filtering, and hallucination-aware variants. Its contribution therefore lies in the evaluation design: converting real videos into object-track grounded, shortcut-controlled, and hallucination-aware diagnostic QA items. Since TOC-Bench evaluates this capability through deterministic QA rather than direct inspection of model internals, its results should be interpreted as behavioral evidence of whether object-level temporal evidence is available to the model. Additional dataset analysis is provided in \cref{appendix:statistics}.

\section{Experiments}
\label{sec:experiments}


\subsection{Evaluation Setup}
We evaluate 23 prominent models on TOC-Bench, covering closed-source multimodal systems, open-source standard Video-LLMs, and open-source thinking or reasoning-style variants. Models are evaluated under a controlled frame-based protocol, with the exact number of sampled frames reported in \cref{tab:tocbench-main-results}; most models use 32 uniformly sampled frames, while several legacy or model-specific baselines follow their feasible input settings. This protocol provides a reproducible visual evidence budget and reduces the influence of hidden API-specific video decoding or internal frame-selection strategies. Additional protocol details are provided in \cref{appendix:frame-protocol}.

For closed-source models, we conduct evaluations via official APIs. For open-source models, all experiments are performed using two A800 GPUs with 80GB memory each. Human verification is used only to filter ambiguous or incorrectly labeled items during benchmark construction, while the human baseline is collected separately on the finalized QA items using only the video and question, without access to object tracks or construction metadata.

We use deterministic scoring for all task formats. Multiple-choice and statement-pair questions are scored by exact label match; numerical questions are normalized to the controlled answer set; and ordering questions require an exact order match.
Beyond overall accuracy and per-dimension accuracy across the 10 diagnostic dimensions, we report hallucination diagnostic accuracy (HDA) on hallucination-aware variants. Let \(\mathcal{H}_A\) and \(\mathcal{H}_B\) denote the nonexistent-subject and nonexistent-event subsets, respectively:
\begin{equation}
\mathrm{HDA}
=
\frac{
\sum_{q\in \mathcal{H}_A \cup \mathcal{H}_B}
\mathbf{1}[\hat{a}_q=a_q]
}{
|\mathcal{H}_A \cup \mathcal{H}_B|
},
\label{eq:hda}
\end{equation}
where \(\hat{a}_q\) and \(a_q\) are the model prediction and ground-truth answer for question \(q\), and \(\mathbf{1}[\cdot]\) is the indicator function. A higher HDA indicates better rejection of plausible but visually unsupported objects or events.

\subsection{Main Results}

\begin{table*}[t]
\caption{
Main evaluation results on TOC-Bench.
All values are percentages.
Frames denotes the number of sampled frames used as input.
Random denotes the expected accuracy of uniform guessing within the valid answer space of each item.
Exist = event existence, Count = event count, ReID = reappearance identity, R/D = reappearance / disappearance, X-Order = cross-object order, Order = event ordering, T-Loc = temporal location, Dur = duration category, Cond = conditional state, and Move = relative spatial change.
HDA denotes hallucination diagnostic accuracy, defined in \cref{eq:hda}.
}
\centering
\scriptsize
\setlength{\tabcolsep}{2.4pt}
\resizebox{\textwidth}{!}{
\begin{tabular}{lccccccccccccc}
\toprule
Model & Frames & Overall & Exist & Count & ReID & R/D & X-Order & Order & T-Loc & Dur & Cond & Move & HDA \\
\midrule
Human & - & 89.3 & 94.5 & 84.0 & 100.0 & 75.4 & 87.2 & 80.0 & 94.5 & 89.7 & 91.4 & 87.6 & 89.5 \\
Random & - & 30.4 & 50.0 & 25.0 & 50.0 & 50.0 & 50.0 & 10.4 & 25.0 & 25.0 & 25.0 & 25.0 & 25.0 \\
\midrule
\multicolumn{14}{l}{\textit{Closed-source models}} \\
GPT-5.5~\cite{openai2026gpt55docs} & 32 & 47.1 & 68.9 & 18.3 & 91.3 & 78.4 & 63.1 & 24.8 & 48.6 & 44.6 & 41.7 & 43.8 & 42.8 \\
Kimi-K2.6 \cite{moonshot2026kimik26modelcard} & 32 & 45.0 & 56.6 & 19.3 & 87.0 & 69.4 & 63.6 & 22.9 & 47.4 & 30.1 & 56.5 & 43.8 & 52.9 \\
Gemini-3.1-Pro-Preview \cite{google2026gemini31propreview} & 32 & 43.9 & 56.2 & 18.7 & 87.0 & 67.2 & 57.2 & 17.1 & 45.0 & 40.1 & 50.0 & 43.8 & 47.7 \\
Grok-4.3 \cite{xai2026grok43} & 32 & 43.8 & 48.9 & 19.7 & 95.7 & 70.9 & 61.0 & 17.1 & 42.9 & 39.7 & 51.8 & 46.6 & 53.8 \\
Seed-2.0-Lite \cite{bytedance2026seed20lite} & 32 & 42.0 & 57.5 & 24.3 & 95.7 & 49.3 & 52.9 & 21.9 & 40.9 & 43.9 & 43.5 & 38.2 & 41.2 \\
GLM-5V-Turbo \cite{zai2026glm5vturbo} & 32 & 37.9 & 52.1 & 13.7 & 60.9 & 67.9 & 56.1 & 17.1 & 34.8 & 42.3 & 36.7 & 33.1 & 34.7 \\
Gemini-3.1-Flash-Lite-Preview & 32 & 33.8 & 37.9 & 6.0 & 91.3 & 58.2 & 48.1 & 16.2 & 38.0 & 30.1 & 39.6 & 28.7 & 41.2 \\
GPT-5.4-mini & 32 & 33.1 & 46.1 & 21.3 & 69.6 & 47.8 & 49.2 & 11.4 & 34.1 & 21.8 & 36.3 & 28.7 & 28.4 \\
Mimo-V2-Omni \cite{xiaomi2026mimov2omni} & 32 & 31.6 & 47.5 & 4.7 & 69.6 & 57.5 & 47.6 & 7.6 & 30.3 & 26.9 & 42.1 & 27.0 & 35.1 \\
\midrule
\multicolumn{14}{l}{\textit{Open-source thinking / reasoning models}} \\
Qwen3-VL-8B-Thinking \cite{bai2025qwen3} & 32 & 32.1 & 35.6 & 14.0 & 100.0 & 35.1 & 50.8 & 10.5 & 34.9 & 24.7 & 40.6 & 30.3 & 43.8 \\
VideoChat-R1.5-7B \cite{yan2025videochat} & 32 & 29.4 & 45.2 & 7.0 & 100.0 & 47.0 & 47.1 & 9.5 & 29.3 & 17.0 & 40.6 & 23.6 & 33.1 \\
Video-R1-7B \cite{feng2025video} & 32 & 25.1 & 34.2 & 0.0 & 100.0 & 32.8 & 46.5 & 14.3 & 25.6 & 15.1 & 35.3 & 24.7 & 32.5 \\
\midrule
\multicolumn{14}{l}{\textit{Open-source standard models}} \\
Qwen2.5-VL-72B \cite{Qwen2.5-VL} & 32 & 34.0 & 47.5 & 3.0 & 95.7 & 49.3 & 50.3 & 15.2 & 38.2 & 30.1 & 40.6 & 27.0 & 42.6 \\
InternVL3-8B \cite{zhu2025internvl3} & 32 & 30.1 & 42.5 & 18.7 & 91.3 & 42.5 & 43.3 & 5.7 & 32.0 & 20.2 & 31.7 & 26.4 & 27.8 \\
LLaVA-Video-72B \cite{zhang2024llava} & 32 & 28.9 & 49.3 & 15.0 & 26.1 & 47.0 & 45.5 & 14.3 & 25.9 & 21.5 & 31.7 & 24.2 & 24.5 \\
VideoLLaMA3-7B \cite{zhang2025videollama} & 32 & 28.4 & 36.1 & 33.7 & 78.3 & 30.6 & 46.0 & 16.2 & 26.4 & 17.0 & 22.3 & 27.0 & 24.7 \\
LLaVA-Video-7B & 8 & 27.9 & 32.0 & 26.3 & 100.0 & 31.3 & 49.2 & 9.5 & 28.6 & 21.5 & 18.0 & 27.0 & 21.7 \\
Video-LLaVA-7B \cite{lin2024video} & 8 & 27.6 & 51.1 & 4.7 & 8.7 & 59.7 & 47.1 & 3.8 & 25.6 & 23.7 & 28.4 & 21.3 & 25.6 \\
MiniCPM-V-2.6 \cite{yao2024minicpm} & 32 & 27.1 & 35.6 & 18.7 & 100.0 & 44.0 & 52.4 & 9.5 & 23.5 & 17.0 & 25.2 & 24.7 & 22.3 \\
Qwen2.5-VL-7B & 32 & 27.0 & 40.6 & 0.7 & 95.7 & 38.8 & 43.3 & 14.3 & 25.9 & 15.1 & 43.9 & 25.8 & 35.3 \\
LLaVA-OV-7B-1Frame \cite{li2024llava} & 1 & 25.4 & 25.1 & 28.3 & 87.0 & 36.6 & 42.8 & 9.5 & 25.7 & 17.9 & 15.8 & 23.0 & 27.8 \\
Qwen3-VL-8B & 32 & 25.2 & 35.6 & 12.3 & 95.7 & 41.8 & 44.9 & 11.4 & 20.1 & 17.6 & 31.3 & 20.2 & 25.4 \\
Video-ChatGPT-7B \cite{maaz2024video} & 100 & 21.0 & 33.3 & 30.7 & 0.0 & 47.8 & 1.1 & 7.6 & 18.1 & 19.6 & 20.5 & 14.0 & 12.4 \\
\bottomrule
\end{tabular}
}
\label{tab:tocbench-main-results}
\vspace{-15pt}
\end{table*}

\begin{figure*}[t]
    \centering
    \includegraphics[width=0.9\textwidth]{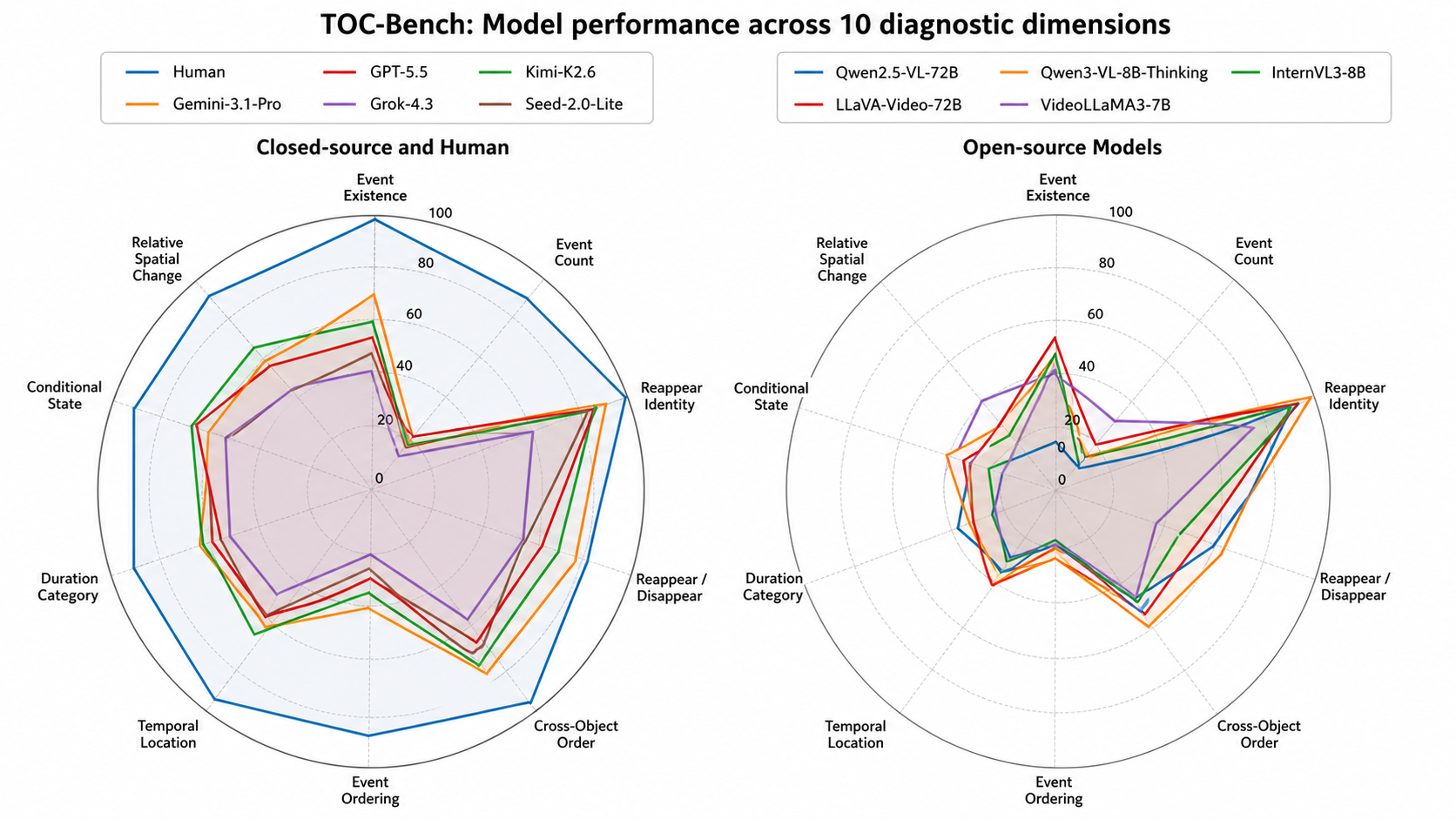}
    \vspace{-5px}
    \caption{
    Radar visualization of model performance across the 10 diagnostic dimensions of TOC-Bench.
    The left plot compares human performance with representative closed-source models, while the right plot compares representative open-source models.
    The visualization highlights the human--model gap and shows that event counting and event ordering remain difficult across model families.
    }
    \vspace{-15px}
    \label{fig:main-result-radar}
\end{figure*}

\paragraph{TOC-Bench reveals a large human--model gap.}
As shown in \cref{tab:tocbench-main-results} and \cref{fig:main-result-radar}, human annotators achieve 89.3\% overall accuracy, while the best-performing model, GPT-5.5, reaches 47.1\%. Kimi-K2.6, Gemini-3.1-Pro-Preview, and Grok-4.3 follow with 45.0\%, 43.9\%, and 43.8\%, respectively. This large gap suggests that TOC-Bench goes beyond surface-level video recognition and exposes failures in maintaining object identity, state, and temporal relations over time.

\paragraph{Closed-source models lead, but temporal object consistency remains unsolved.}
Closed-source models achieve the strongest overall results, averaging 39.8\% accuracy. The best open-source standard model, Qwen2.5-VL-72B, reaches 34.0\%, while the best open-source thinking model, Qwen3-VL-8B-Thinking, reaches 32.1\%. Model scale and proprietary training improve performance, but the gap to human accuracy remains large, supporting our main claim that temporal object consistency is not well captured by existing Video-LLMs. Since Qwen3-VL-8B-Instruct is used as the filtering model, the retained set may be relatively harder for Qwen-style models under shortcut settings; we discuss this dependence in \cref{appendix:filter-model}.

\paragraph{Thinking-style models help selectively but are not a complete solution.}
Thinking or reasoning-style variants do not consistently outperform standard models. Qwen3-VL-8B-Thinking improves over Qwen3-VL-8B in both overall accuracy and HDA, but VideoChat-R1.5-7B and Video-R1-7B remain below several standard models. This suggests that reasoning-style inference can help in some cases, but it cannot fully compensate for missing object-level temporal evidence. When the visual representation does not reliably preserve object identity, repeated events, or frame-order information, longer reasoning traces alone are insufficient to solve temporal object consistency.



\paragraph{Event counting, ordering, and hallucination verification are key failure modes.}
Models struggle most with structured temporal accumulation and chronological reconstruction. For example, GPT-5.5 reaches 68.9\% on event existence and 78.4\% on reappearance/disappearance, but only 18.3\% on event counting and 24.8\% on event ordering. HDA further reveals a distinct robustness axis: GPT-5.5 has the best overall score, while Grok-4.3 and Kimi-K2.6 achieve higher HDA scores. Thus, a model may answer ordinary temporal questions correctly while still over-assuming plausible but unsupported objects or events.

\paragraph{Implications for model design.}
The failure patterns in TOC-Bench suggest that temporal object consistency is not solved by stronger language reasoning alone. Many Video-LLMs represent videos through sparse frame sampling, visual-token compression, or global temporal aggregation. These mechanisms are effective for coarse event recognition, but can discard the frame-level continuity needed to count repeated object events, distinguish occlusion from leaving the frame, and maintain identity across reappearance. In addition, much video-instruction data is organized around captions, general QA, or video-level events rather than object-track grounded supervision. This may explain why reasoning-style models improve some metrics but do not close the gap: if object-level temporal evidence is not preserved in the visual representation, longer reasoning traces cannot reliably recover it. Future models may benefit from object-centric memory, tracking-aware visual tokens, explicit visibility and event supervision, and training objectives that model object identity, event timestamps, and cross-object temporal relations.

Additional evaluation breakdowns, including tier-level accuracy, format-level accuracy, hallucination-bucket results, and model-by-dimension visualizations, are provided in \cref{appendix:additional-results}.

\section{Conclusion}
\label{sec:conclusion}

We introduce TOC-Bench, a diagnostic benchmark for evaluating temporal object consistency in Video-LLMs. By grounding questions in object tracks and event timelines, TOC-Bench tests whether models can maintain object identity, state, and persistence across occlusion, reappearance, and cross-object interactions. It further combines temporal-necessity filtering with hallucination-aware variants to reduce shortcut-solvable questions and probe unsupported object/event assumptions.
Experiments show that current Video-LLMs remain far below human performance, especially on event counting, event ordering, identity-sensitive reasoning, and hallucination-aware verification. This suggests that strong general video understanding does not yet imply reliable object-level temporal reasoning. The current verified set prioritizes diagnostic quality over scale; future work can expand TOC-Bench with larger evaluation and training splits, and use this object-track grounded paradigm to train models with stronger object-centric memory and temporally grounded reasoning.

\newpage


\bibliographystyle{plain}
\bibliography{refs}

\clearpage

\appendix
\section{More Details about Event Detection}
\label{appendix:event-detection}

\subsection{Noun-Phrase Extraction}

For each source video, we first sample frames at 1 fps and limit the input to a maximum of 10 frames. The sampled frames are passed to Qwen3-VL-8B-Instruct in a single multimodal turn to obtain a list of object-centric noun phrases. We use a VLM rather than a fixed object vocabulary because TOC-Bench draws from diverse video sources, where relevant objects may vary substantially across scenes.

The prompt asks the model to enumerate distinct physical objects and people, encourages discriminative attributes such as color, material, clothing, or action, and explicitly excludes background entities. The exact system prompt is shown in \cref{fig:vlm-prompt}:

\begin{figure}[t]
    \centering
    \includegraphics[width=1.0\linewidth]{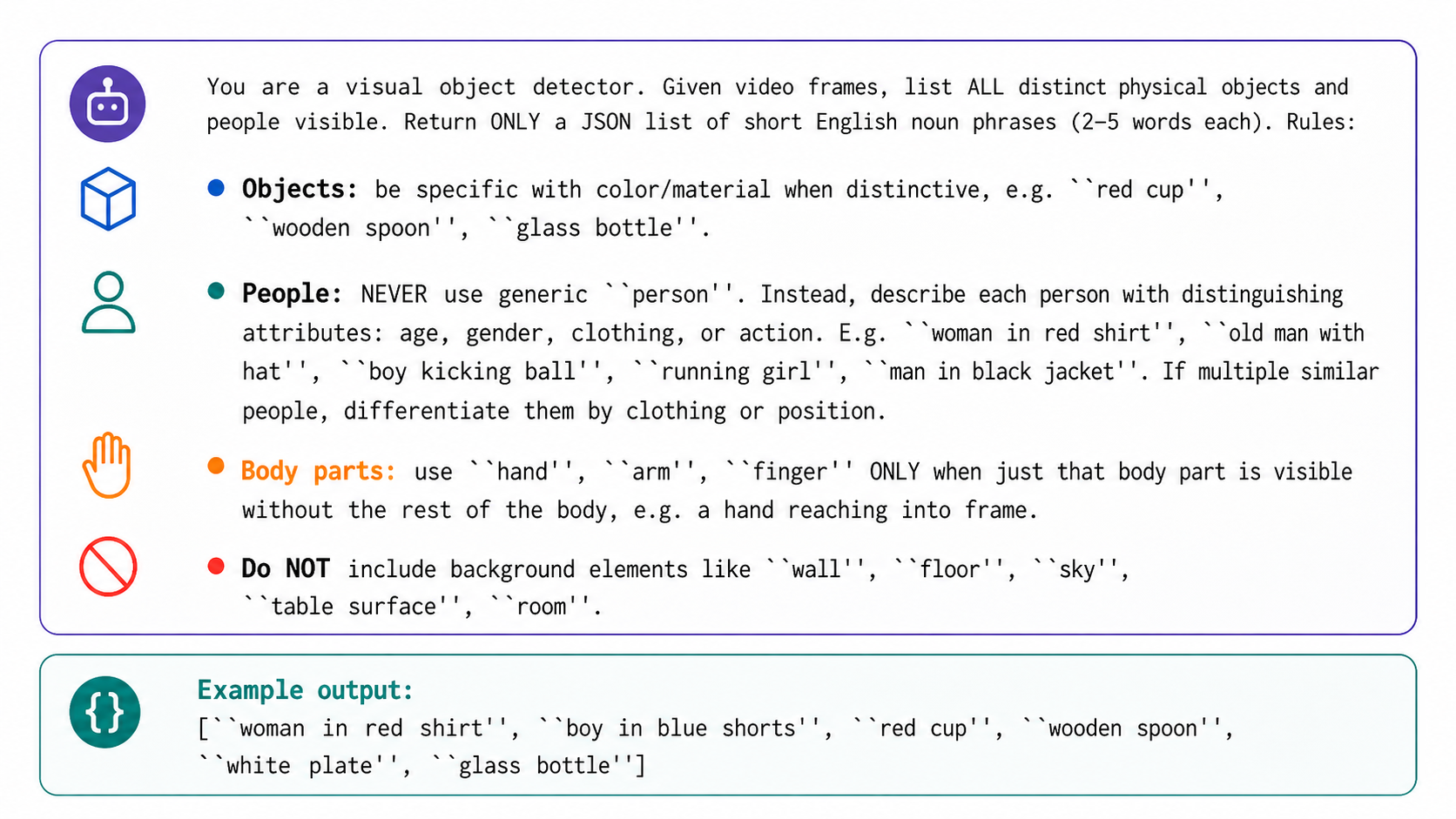}
    \caption{The exact system prompt for VLM.}
    \label{fig:vlm-prompt}
\end{figure}

We keep up to 15 noun phrases per video. If parsing fails or the model returns an empty list, we use a fixed fallback prompt pool covering common categories: \texttt{person}, \texttt{hand}, \texttt{cup}, \texttt{bottle}, \texttt{ball}, \texttt{box}, \texttt{bag}, \texttt{animal}, \texttt{vehicle}, and \texttt{food}. The resulting noun phrases are cached and used as text prompts for the tracking stage.

\subsection{SAM3 Tracking}

We use the extracted noun phrases as text prompts for SAM3 in video-predictor mode. Each noun phrase is independently fed into SAM3, which produces per-frame masks, bounding boxes, confidence scores, and visibility flags. Tracking is performed at 3 fps. We retain masks that satisfy the confidence and area thresholds, and limit the number of tracks per video to avoid excessive noisy instances.

For each video, the raw tracking output is stored as:
\begin{equation}
\mathcal{O}
=
\{(o_i, \mathbf{m}_i^{1:T}, \text{label}_i)\}_{i=1}^{K},
\label{eq:raw-tracking-output}
\end{equation}
where \(\mathcal{O}\) is the set of raw object tracks, \(K\) is the number of tracked objects, \(T\) is the number of tracked frames, \(o_i\) is the object ID, \(\mathbf{m}_i^{1:T}\) is the per-frame binary mask sequence, and \(\text{label}_i\) is the noun phrase used to obtain the track. Each track also stores per-frame bounding boxes, visibility flags, confidence scores, and mask areas.

\subsection{Track Post-processing}

Raw SAM3 tracks may contain unstable short fragments, ID splits, and ID swaps, especially in videos with occlusion or visually similar objects. We therefore apply a lightweight post-processing step before event detection.

First, short fragments are removed if they are visible for too few frames. Second, possible ID splits are repaired by merging same-label tracks whose visible time windows do not overlap and whose positions are spatially continuous. Third, possible ID swaps are flagged when a track exhibits an abrupt position jump between nearby visible frames. We do not automatically split such tracks; instead, the detected swaps are recorded and used to reduce the track-level confidence score.

Each retained track is assigned an instance confidence score:
\begin{equation}
c
=
c_{\text{base}}
\cdot
\pi_{\text{split}}
\cdot
\pi_{\text{swap}}
\cdot
\beta_{\text{cov}},
\label{eq:instance-confidence-score}
\end{equation}
where \(c_{\text{base}}\) is the mean SAM3 confidence over visible frames, \(\pi_{\text{split}}\) penalizes merge operations, \(\pi_{\text{swap}}\) penalizes unresolved swap detections, and \(\beta_{\text{cov}}\) reflects temporal coverage. Tier-2 and Tier-3 reasoning units use this score to filter unreliable subject tracks, so that identity-sensitive questions are not generated from unstable object trajectories.

\subsection{Temporal Event Detection}

After track post-processing, we extract temporal events from each object's mask and bounding-box sequence. The detector uses per-frame visibility, mask-area changes, object positions, and pairwise bounding-box overlap to identify object-level and pairwise events.

We detect eight per-object event types: \texttt{appear}, \texttt{enter\_frame}, \texttt{exit\_frame}, \texttt{partial\_occlusion}, \texttt{full\_occlusion}, \texttt{reappear}, \texttt{disappear}, and \texttt{state\_change}. We also detect one pairwise event type, \texttt{interaction}, when two object tracks overlap for a sustained period. For occlusion events, we further store possible occluder candidates by checking nearby objects whose bounding boxes overlap with the target object around the occlusion onset.

Each event is stored with metadata needed for QA construction, including its object ID, event type, frame or frame range, timestamp, temporal-position bucket, duration bucket when applicable, and possible related objects. We also build object timelines by sorting each object's events chronologically. These timelines are later used to construct event-ordering, reappearance, event-count, and cross-object reasoning units.

\subsection{Video-Level Retention}

Finally, we filter videos based on whether they contain enough object-level temporal structure for QA construction. A video is retained only if it has sufficient duration, at least two tracked objects, enough meaningful temporal events, and either occlusion-related or interaction-related phenomena. Each retained video is also assigned a quality score that favors richer object-temporal dynamics, such as occlusion, reappearance, interaction, and multi-object involvement. This score is later used when sampling videos for reasoning-unit construction. Main hyperparameters used in Stage 1 are shown in \cref{tab:stage1-hyperparams}.

\begin{table}[ht]
\caption{Main hyperparameters used in Stage 1 object-temporal annotation.}
\centering
\small
\begin{tabular}{lll}
\toprule
Component & Parameter & Value \\
\midrule
Noun-phrase extraction & Frame sampling rate & 1 fps \\
Noun-phrase extraction & Maximum frames per video & 10 \\
Noun-phrase extraction & Maximum noun phrases per video & 15 \\
Noun-phrase extraction & Fallback prompt pool size & 10 \\
\midrule
SAM3 tracking & Tracking frame rate & 3 fps \\
SAM3 tracking & Confidence threshold & 0.3 \\
SAM3 tracking & Minimum mask area & 100 px \\
SAM3 tracking & Maximum tracks per video & 20 \\
\midrule
Track post-processing & Minimum fragment length & 5 frames or 2\% of video \\
Track post-processing & Maximum split-repair time gap & 2 s \\
Track post-processing & Maximum split-repair spatial jump & 15\% of image diagonal \\
Track post-processing & Swap-detection spatial jump & 35\% of image diagonal \\
Track post-processing & Swap-detection visible-frame gap & 3 frames \\
Track post-processing & Tier-2/Tier-3 confidence cutoff & 0.7 \\
\midrule
Event detection & Partial-occlusion area ratio & 0.50 \\
Event detection & Full-occlusion area ratio & 0.10 \\
Event detection & Minimum occlusion segment length & 2 frames \\
Event detection & Reappearance confirmation & 2 visible frames \\
Event detection & Interaction IoU threshold & 0.05 \\
Event detection & Minimum interaction length & 3 frames \\
\midrule
Video retention & Duration range & 5--90 s \\
Video retention & Minimum tracked objects & 2 \\
Video retention & Minimum meaningful events & 3 \\
Video retention & Minimum interactions if no occlusion & 2 \\
\bottomrule
\end{tabular}
\label{tab:stage1-hyperparams}
\end{table}
\section{More Details about QA Construction}
\label{appendix:qa-construction}

This appendix provides implementation details for Stage 2, which converts the object-level temporal annotations from Stage 1 into candidate question-answer pairs. The stage consists of four steps: video sampling, reasoning-unit construction, skeleton synthesis with hallucination injection, and surface realization. A summary of the main implementation settings is provided in \cref{tab:stage2-implementation}.

\subsection{Video Sampling}
\label{app:stage2-video-sampling}

We select 6,000 videos from the Stage-1 retained pool before constructing reasoning units. Since the retained videos are not uniformly distributed over object-level temporal phenomena, we use a phenomenon-first sampling strategy. The goal is to ensure that the selected videos cover both common temporal events, such as appearance and interaction, and rarer consistency-critical phenomena, such as reappearance, repeated events, cross-event chains, and high-confidence multi-object cases.

For each candidate video \(v\), we compute a sampling score:
\begin{equation}
\begin{aligned}
S(v)
=&
\underbrace{
\sum_{s\in\mathcal{S}}
w_s \cdot \mathbf{1}[v\text{ covers }s]\cdot \Delta_s
}_{\text{phenomenon coverage}}
\\
&+
\underbrace{
\lambda_{\mathrm{dur}}
\Delta_{\mathrm{dur}}(b_{\mathrm{dur}}(v))
}_{\text{duration balance}}
+
\underbrace{
\lambda_{\mathrm{den}}
\Delta_{\mathrm{den}}(b_{\mathrm{den}}(v))
}_{\text{object-density balance}}
\\
&-
\underbrace{
\lambda_{\mathrm{src}}\mathbf{1}[\text{source}(v)\text{ exceeds cap}]
}_{\text{source penalty}} .
\end{aligned}
\label{eq:candidate-video-score}
\end{equation}

Here, \(\mathcal{S}\) denotes the set of phenomenon slots, \(w_s\) is the slot weight, and \(\mathbf{1}[v\text{ covers }s]\) indicates whether video \(v\) contains phenomenon slot \(s\). The term \(\Delta_s=\max(0,\tau_s-r_s)\) denotes the remaining deficit of slot \(s\), where \(\tau_s\) is the target ratio and \(r_s\) is the current selected ratio. Similarly, \(b_{\mathrm{dur}}(v)\) and \(b_{\mathrm{den}}(v)\) denote the duration bucket and object-density bucket of video \(v\), and \(\Delta_{\mathrm{dur}}\) and \(\Delta_{\mathrm{den}}\) measure the remaining deficits of those buckets. The source penalty discourages selecting videos from a source whose current share has exceeded its cap, with \(\lambda_{\mathrm{src}}>0\). Once a phenomenon or bucket reaches its target ratio, its deficit becomes zero and no longer increases the score. If all major phenomenon deficits are filled before reaching the target number of videos, the remaining videos are selected by their Stage-1 quality scores.

\subsection{Reasoning Unit Construction}
\label{app:stage2-units}

For each selected video, we convert its object timelines and temporal events into structured reasoning units. A reasoning unit is an intermediate representation of a possible QA item:
\begin{equation}
u=(\mathrm{dim}, \mathrm{video\_id}, \mathrm{subject}, \mathrm{payload}),
\label{eq:reasoning-unit}
\end{equation}
where \(\mathrm{dim}\) specifies the diagnostic dimension, \(\mathrm{subject}\) identifies the queried object track, and \(\mathrm{payload}\) stores dimension-specific information such as the anchor event, correct temporal bucket, duration category, event count, sibling object, partner object, or correct ordering relation.

Reasoning units are generated by dimension-specific builders. Each builder enforces hard preconditions before emitting a unit. For example, temporal-location units are generated only when the relevant event occurrence is not ambiguous; duration-category units require a well-defined ranged event; event-count units require repeated events of the same type; and cross-object order units require two temporally separated events from different object tracks. For Tier-2 and Tier-3 dimensions, the involved object tracks must pass the instance-confidence gate introduced in Stage 1. This prevents identity-sensitive questions from being built on unstable tracks.

Across the selected 6,000 videos, this step produces 547,610 candidate reasoning units across 10 diagnostic dimensions. These units are not yet natural-language questions. Instead, they provide the structured logical basis for skeleton construction.

\subsection{Question Skeleton Synthesis}
\label{app:stage2-skeletons}

The reasoning units are then rendered into question skeletons using dimension-specific templates. A skeleton is a fully specified QA record containing a question stem, answer format, correct answer, and format-specific options or statements. The correct answer is determined entirely from the structured reasoning unit, not by a language model.

TOC-Bench uses four task types: multiple-choice, statement pair, numerical, and ordering. Multiple-choice skeletons are used for bucket-style dimensions such as temporal location, duration category, relative spatial change, and conditional state. Statement-pair skeletons are used for binary judgments such as event existence, reappearance/disappearance mechanism, reappearance identity, and cross-object order. Numerical skeletons are used for event-count questions, while ordering skeletons ask the model to arrange three or four events chronologically.

Distractors are generated in a dimension-aware way. Bucket-style dimensions sample distractors from the corresponding bucket vocabulary. Statement-pair dimensions construct the contrastive statement by reversing the mechanism, identity, polarity, or event order. Numerical questions use the controlled answer set \(\{2,3,4,\text{``5 or more''}\}\). Ordering questions present events in shuffled order while storing the correct chronological sequence.

\subsection{Hallucination-aware Variant Construction}
\label{app:stage2-hallucination}

To test whether models over-assume the existence of objects or events, we inject hallucination-aware variants during skeleton construction. Hallucination injection is applied to selected multiple-choice dimensions, including temporal location, duration category, relative spatial change, and conditional state.

We use three treatments. In variant A, the subject in the question is replaced with a plausible but nonexistent object, and the correct answer becomes a statement of the form ``there is no \(X\) in this video''. In variant B, the subject exists, but the described event does not occur, and the correct answer becomes a statement of the form ``\(X\) never \(Y\)''. In the distractor-only treatment, the original question remains valid, but one wrong option is replaced with a hallucination-style distractor. The relative-spatial-change dimension excludes variant B because ``never moves'' would conflict with the valid answer ``stays in roughly the same place''.

Hallucinated subjects are generated using two strategies. Most are created by same-base-noun synthesis, where the base noun is kept but the modifier is replaced by a different color, material, or descriptive attribute. A smaller portion is generated by sampling a phrase from the global label pool whose base noun does not appear in the current video. In both cases, the generated subject is deduplicated against the video's actual object labels and the real subject label.

After hallucination injection, we apply answer balancing and a per-video cap. Answer balancing down-samples each dimension so that correct-answer labels are less biased, while keeping hallucination-tagged skeletons separate from normal answer buckets. The per-video cap prevents a small number of videos from dominating the QA pool and uses round-robin selection across dimensions to preserve diversity. After these steps, 45,527 skeletons are retained for surface realization.

\subsection{Dimension-aware Surface Realization}
\label{app:stage2-surface-realization}

Finally, skeletons are converted into fluent natural-language QA items using GPT-5.4-mini. The LLM is only used to polish the surface form. It does not determine the correct answer, modify the answer label, or change any structured metadata.

\begin{figure}[t]
    \centering
    \includegraphics[width=1.0\linewidth]{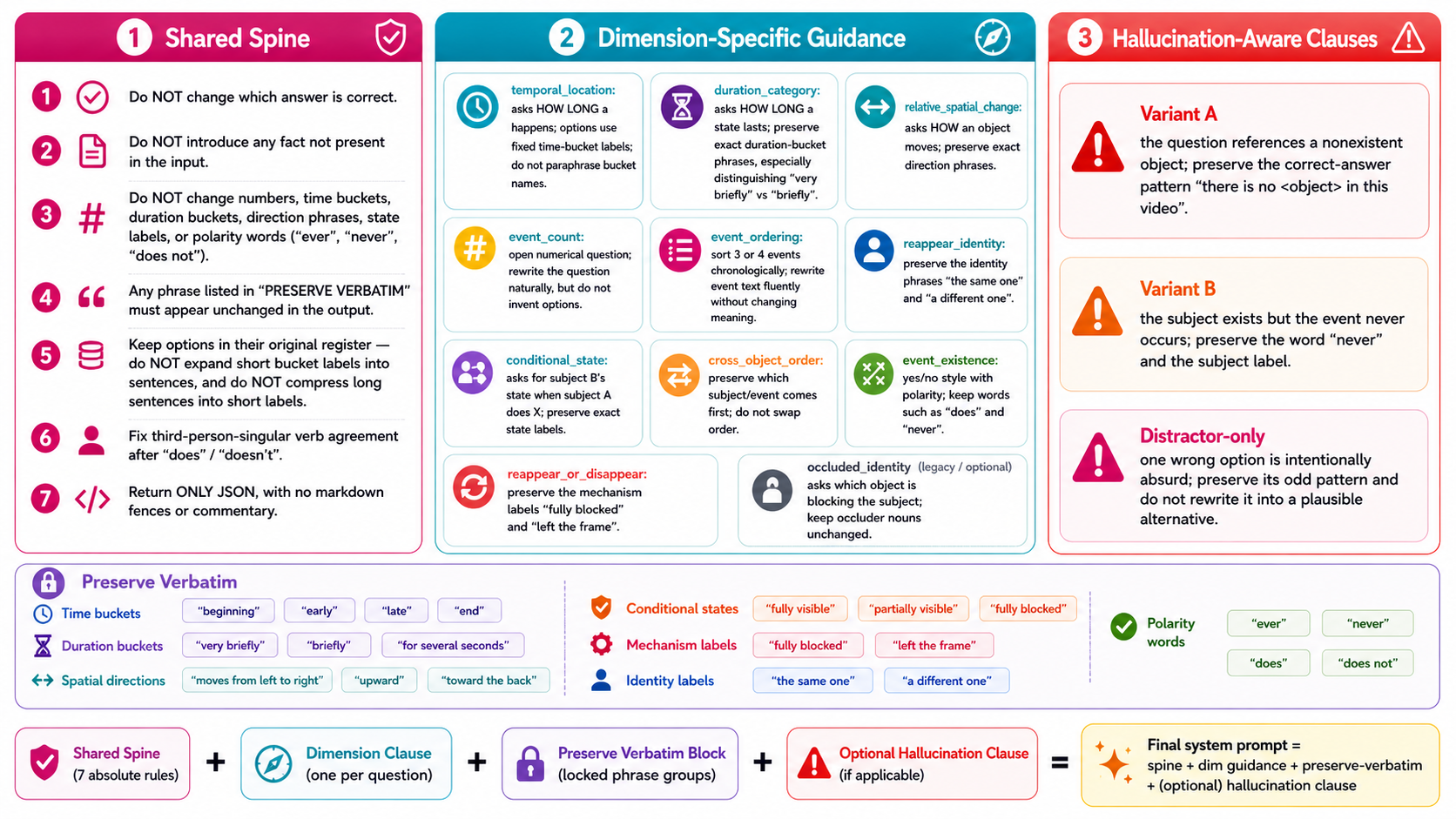}
    \caption{
    Dimension-aware surface-realization prompt design used in Stage 2.
    The system prompt consists of a shared instruction spine, a dimension-specific clause, a preserve-verbatim block for locked labels, and an optional hallucination-aware clause.
    This design allows GPT-5.4-mini to improve linguistic fluency while preserving deterministic answer semantics.
    }
    \label{fig:stage2-surface-realization-prompt}
\end{figure}

To reduce semantic drift, each call uses a dimension-aware system prompt. As illustrated in \cref{fig:stage2-surface-realization-prompt}, the final prompt is assembled from four components: a shared instruction spine, a dimension-specific guidance clause, a preserve-verbatim block, and an optional hallucination-specific clause. The shared spine imposes global constraints, including not changing the correct answer, not introducing unsupported facts, preserving numbers and controlled labels, keeping option registers unchanged, fixing minor grammatical issues, and returning only JSON.

The dimension-specific clause explains what each dimension is asking and which parts must remain unchanged. For example, temporal-location items instruct the model that the options are fixed time-bucket labels; duration-category items require preserving the distinction between ``very briefly'' and ``briefly''; relative-spatial-change items require preserving direction phrases; conditional-state items require preserving visibility-state labels; and ordering items instruct the model to rewrite event descriptions without changing their chronological semantics.

For dimensions with controlled answer vocabularies, the prompt also includes a \texttt{PRESERVE VERBATIM} block. This block lists phrases that must survive rewriting exactly. The locked phrases include temporal buckets, duration buckets, spatial-direction labels, conditional-state labels, reappearance identity labels, mechanism labels, and key polarity words. The main locked vocabularies are summarized in \cref{tab:surface-realization-locks}.

\begin{table}[t]
\caption{
Representative locked phrases used during dimension-aware surface realization.
These phrases must be preserved verbatim in the polished QA item.
}
\centering
\small
\begin{tabularx}{\linewidth}{@{}l>{\raggedright\arraybackslash}X@{}}
\toprule
Dimension & Locked phrases \\
\midrule
\texttt{temporal\_location} &
\texttt{beginning}, \texttt{early}, \texttt{late}, \texttt{end} \\

\texttt{duration\_category} &
\texttt{very briefly}, \texttt{briefly}, \texttt{for several seconds}, \texttt{for an extended period} \\

\texttt{relative\_spatial\_change} &
\texttt{moves from left to right}, \texttt{moves from right to left}, \texttt{moves from top to bottom}, \texttt{moves from bottom to top}, \texttt{stays in roughly the same place} \\

\texttt{conditional\_state} &
\texttt{fully visible}, \texttt{partly hidden}, \texttt{fully hidden}, \texttt{not yet appeared or already gone} \\

\texttt{reappear\_or\_disappear} &
\texttt{fully blocked}, \texttt{left the frame} \\

\texttt{reappear\_identity} &
\texttt{the same one}, \texttt{a different one} \\
\bottomrule
\end{tabularx}
\label{tab:surface-realization-locks}
\end{table}

For hallucination-related skeletons, an additional clause is appended to the system prompt. For variant-A hallucination questions, where the question references a nonexistent object, the prompt explicitly requires preserving the pattern ``there is no \(\langle\)object\(\rangle\) in this video'' in the correct option. For variant-B hallucination questions, where the subject exists but the queried event never happens, the prompt requires preserving the word ``never'' and the same subject label. For normal questions that contain a hallucination-style distractor, the prompt instructs the model not to rewrite the intentionally absurd option into a plausible alternative.

After surface realization, each item is automatically checked. The verifier detects empty fields, duplicate options, bucket-label violations, hallucination-marker violations, numerical-answer violations, and structural errors in ordering questions. If a polished field violates a locked phrase constraint, that field is rolled back to the original skeleton text. For hallucination-marker violations, all four options are rolled back to avoid changing the intended shortcut-resistant structure. This mechanism ensures that the final QA items are more natural while preserving structured semantics and deterministic answer labels.

\begin{table}[t]
\caption{
Summary of the main implementation settings in Stage 2.
}
\centering
\small
\begin{tabular}{lll}
\toprule
Component & Setting & Value \\
\midrule
Video sampling & Selected videos & 6,000 \\
Video sampling & Duration buckets & short / medium / long \\
Video sampling & Object-density buckets & sparse / moderate / dense \\
Video sampling & Maximum source ratio & 0.45 \\
\midrule
Reasoning units & Total reasoning units & 547,610 \\
Reasoning units & Diagnostic dimensions & 10 \\
Reasoning units & Tier-2/Tier-3 confidence gate & \(c \geq 0.7\) \\
\midrule
Skeletons & Retained skeletons & 45,527 \\
Skeletons & Task types & multiple-choice / statement pair / numerical / ordering \\
Hallucination & Variant A & nonexistent subject \\
Hallucination & Variant B & nonexistent event \\
Hallucination & Distractor-only & hallucination-style wrong option \\
\midrule
Surface realization & Realization model & GPT-5.4-mini \\
Surface realization & Prompt design & shared spine + dim clause + preserve-verbatim block \\
Surface realization & Hallucination protection & preserve ``there is no'' and ``never'' patterns \\
Surface realization & Error handling & field-level rollback \\
\bottomrule
\end{tabular}
\label{tab:stage2-implementation}
\end{table}
\section{More Details about Filtering and Verification}
\label{appendix:filtering}

\subsection{Three-Layer Temporal-Necessity Filter}
\label{app:temporal-necessity-filter}

Given the natural-language QA pool generated in Stage 2, we apply three independent filtering layers. Each layer tests whether a question can be solved without the intended form of temporally ordered object-level reasoning. A QA item is retained only if it passes all applicable layers.

The three layers are:

\paragraph{Layer 1: text-only filter.}
The model receives only the question and answer choices, without any visual input. If Qwen3-VL-8B-Instruct answers the item correctly in this setting, the item is considered solvable from language priors, answer-option bias, or textual leakage, and is removed.

\paragraph{Layer 2: single-frame filter.}
The model receives a single video frame together with the question. If the item can be answered correctly from individual frames often enough, it is considered not to require temporal reasoning and is removed. Instead of using only one random frame, we probe frames progressively from a set of uniformly sampled frames. The probing order is temporally dispersed, so early probes cover different parts of the video. This reduces the chance that a single unlucky frame determines whether an item is kept or removed.

\paragraph{Layer 3: frame-shuffled filter.}
The model is evaluated on the original ordered frame sequence and on a shuffled version of the same frames. If the item is answered correctly under both ordered and shuffled inputs, it is treated as not requiring temporal order and is removed. Items that are solved only with the ordered input are retained. Items that the model fails under both conditions are also retained, since they remain difficult and do not show evidence of being solvable from unordered frame evidence.

\subsection{Format-Aware Answer Parsing}
\label{app:format-aware-filtering}

Because TOC-Bench contains multiple task formats, the filtering model output is parsed in a format-aware manner. For multiple-choice questions, the prediction must be one of \texttt{A}, \texttt{B}, \texttt{C}, or \texttt{D}. For statement-pair questions, the prediction must be \texttt{A} or \texttt{B}. For ordering questions, the prediction is parsed as an ordered sequence of event labels, such as \texttt{C,A,B}. For numerical questions, the prediction is normalized into one of the controlled answers: \texttt{2}, \texttt{3}, \texttt{4}, or \texttt{5 or more}.

This format-aware parsing avoids treating free-form explanations as valid answers and ensures that the filtering decision is deterministic. In particular, numerical answers are strictly matched to the controlled answer set, and ordering answers must match the full chronological label sequence.

\subsection{Layer Combination}
\label{app:filter-layer-combination}

After the three filter layers are run independently, their outputs are combined by intersection. An item enters the filtered pool only if it passes all available layers:
\begin{equation}
\mathrm{Pass}(q)
=
\mathrm{Pass}_{\mathrm{text}}(q)
\wedge
\mathrm{Pass}_{\mathrm{single}}(q)
\wedge
\mathrm{Pass}_{\mathrm{shuffled}}(q).
\label{eq:qa-filter-pass}
\end{equation}
The combined file stores the retained items together with their layer-level results. In our construction, the three-layer filter reduces the generated QA pool from 45,527 items to 17,900 filtered QA pairs. These filtered items serve as the candidate pool for subsequent human verification.

\subsection{Sampling for Human Verification}
\label{app:human-verification-sampling}

After filtering, we sample a subset from the 17,900 filtered QA pairs for human verification. The purpose of this step is to obtain a high-quality evaluation set while preserving diagnostic coverage across dimensions. Sampling is performed at the QA-item level rather than the video level, so different QA items may still come from the same source video.

We use dimension-stratified sampling to preserve the diagnostic structure of TOC-Bench. We retain as many items as possible from smaller dimensions to maintain statistical power, while downsampling larger dimensions with per-dimension caps.  This prevents high-frequency dimensions from dominating the final benchmark and ensures that rare but important Tier-2 and Tier-3 dimensions remain represented.

In total, we sample 3,000 QA items for human verification. Annotators inspect the corresponding videos, questions, answer candidates, and ground-truth labels. Minor issues, such as unclear wording, duplicated options, or slight reference ambiguity, are manually revised when the intended visual evidence and answer remain valid. Items are removed only when they contain severe evidence or label problems, such as tracking errors, insufficient visual support, incorrect event labels, wrong answer labels, or a clear mismatch between the QA item and the video. After human verification, TOC-Bench contains 2,323 high-quality QA pairs.
\section{Model-dependence and Bias Control}
\label{appendix:model-dependence}

TOC-Bench uses foundation models as scalable tools during benchmark construction. To keep the final labels independent of any single construction model, we separate model-assisted proposal generation from ground-truth determination. Upstream models help propose objects, tracks, filters, or surface forms, but the final answer labels are fixed by structured object-event records and retained only after rule-based checks and human verification.

\subsection{Filtering-model dependence}
\label{appendix:filter-model}
Qwen3-VL-8B-Instruct is used as the filtering model in the temporal-necessity filter. This introduces a specific dependence: items that Qwen3-VL-8B-Instruct can solve under text-only, single-frame, or shuffled-frame shortcut settings are removed, so the retained set may be relatively harder for Qwen-style models. We therefore report Qwen3 results for completeness but interpret them with this dependence in mind. This filtering choice does not favor the filtering model: items solved by Qwen3-VL-8B-Instruct under shortcut settings are removed rather than selected.

We use a single strong filtering model rather than a multi-model filtering ensemble for two reasons. First, the filter is intended to remove obvious shortcut-solvable items under a fixed probing protocol, not to define semantic correctness. Using multiple models with different visual encoders, context lengths, prompting styles, and answer biases could make the retained set depend on an unstable union or intersection of heterogeneous model behaviors. Second, aggressive multi-model filtering may remove valid but difficult questions simply because one model succeeds under an idiosyncratic shortcut. We therefore use one consistent filtering model, make this dependence explicit, and rely on rule-based constraints and human verification to determine the final benchmark validity.

\subsection{Construction-model dependence}
Other upstream models are used to improve candidate quality rather than to determine final answers. Qwen3-VL-8B-Instruct proposes object noun phrases, SAM3 produces candidate object tracks, and GPT-5.4-mini performs surface realization from structured question skeletons. These models do not independently decide the ground-truth labels. Each reasoning unit fixes the queried subject, diagnostic dimension, answer type, correct answer, and admissible distractors before natural-language generation. GPT-5.4-mini only improves the wording of a pre-specified QA record, while object references, answer labels, temporal buckets, numerical answers, and ordering relations remain locked by the skeleton.

Similarly, SAM3 tracks are treated as candidate evidence rather than automatically trusted labels. Low-quality, fragmented, or ambiguous tracks are filtered before event derivation, and human verification removes items whose answers depend on tracking errors, ambiguous references, insufficient visual evidence, duplicated options, or unclear wording. Thus, foundation models improve the scalability and quality of candidate construction, while the final benchmark emphasizes structured records, rule-based checks, and human-verified labels.
\section{More Details about Dataset Statistics}
\label{appendix:statistics}

\subsection{Source-Dimension Coverage}

\begin{figure}[t]
    \centering
    \includegraphics[width=0.8\linewidth]{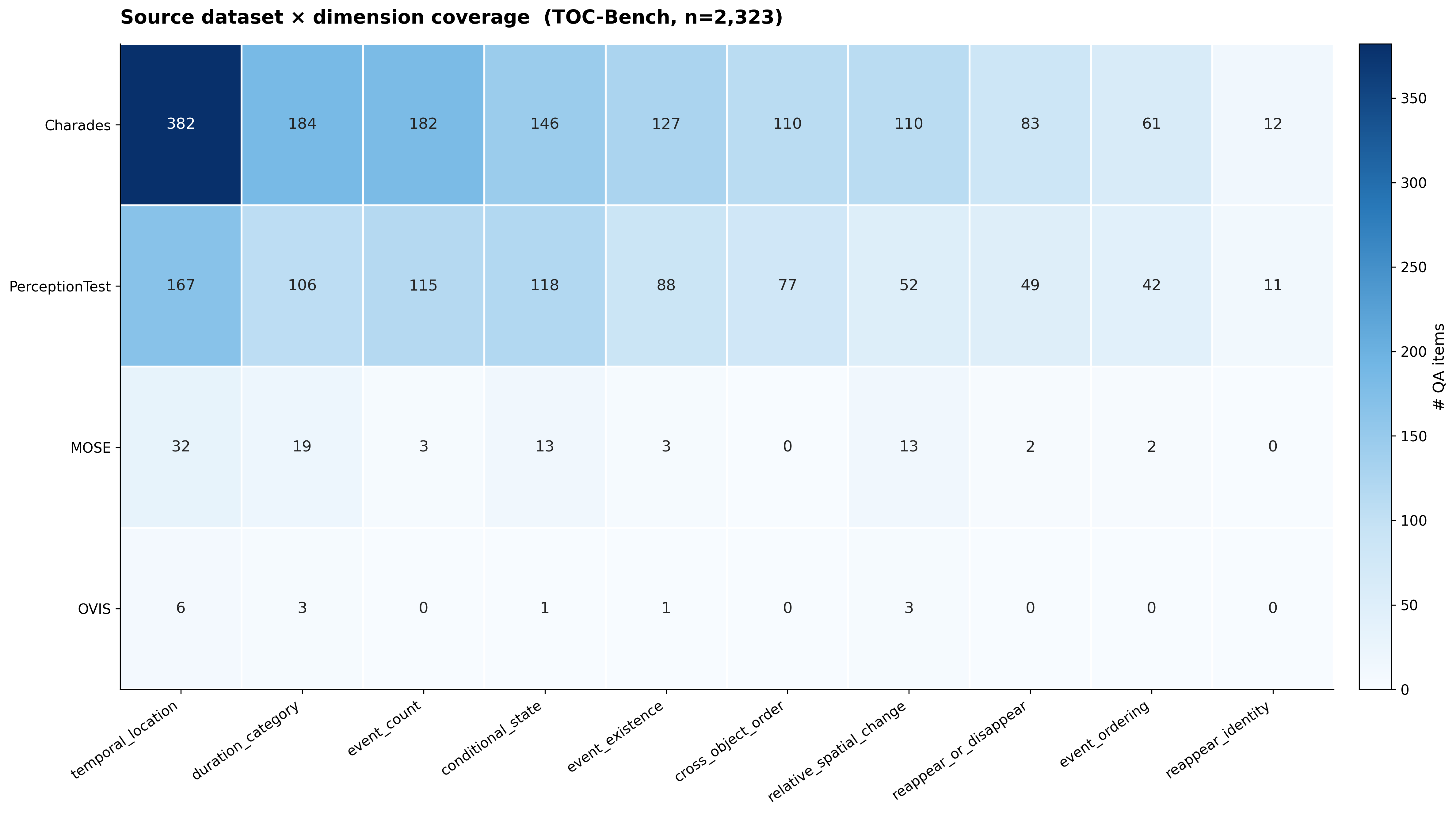}
    \caption{
    Source dataset by diagnostic-dimension coverage in TOC-Bench.
    Charades and Perception Test contribute most final QA items, while MOSE and OVIS provide additional occlusion-heavy cases.
    The heatmap shows that the final benchmark covers multiple diagnostic dimensions across source datasets rather than relying on a single source-dimension pair.
    }
    \label{fig:source-dimension-coverage}
\end{figure}

\Cref{fig:source-dimension-coverage} shows the distribution of diagnostic dimensions across source datasets. Although Charades and Perception Test dominate the final verified set, the retained QA items span multiple dimensions across sources. MOSE and OVIS contribute fewer final items, but their occlusion-heavy videos are useful for constructing and validating rare object-level temporal phenomena during the benchmark construction process.

\subsection{Dimension-wise Qualitative Examples}
\label{appendix:dimension-examples}

To make the diagnostic dimensions more concrete, we provide representative visual QA examples for all 10 dimensions of TOC-Bench. Each example shows the dimension, tier, task format, natural-language question, sampled video frames with timestamps, and answer candidates with the correct answer highlighted. These examples illustrate how TOC-Bench grounds questions in object-level temporal evidence rather than relying only on video-level captions or free-form object mentions.

\begin{figure*}[t]
    \centering
    \includegraphics[width=\textwidth]{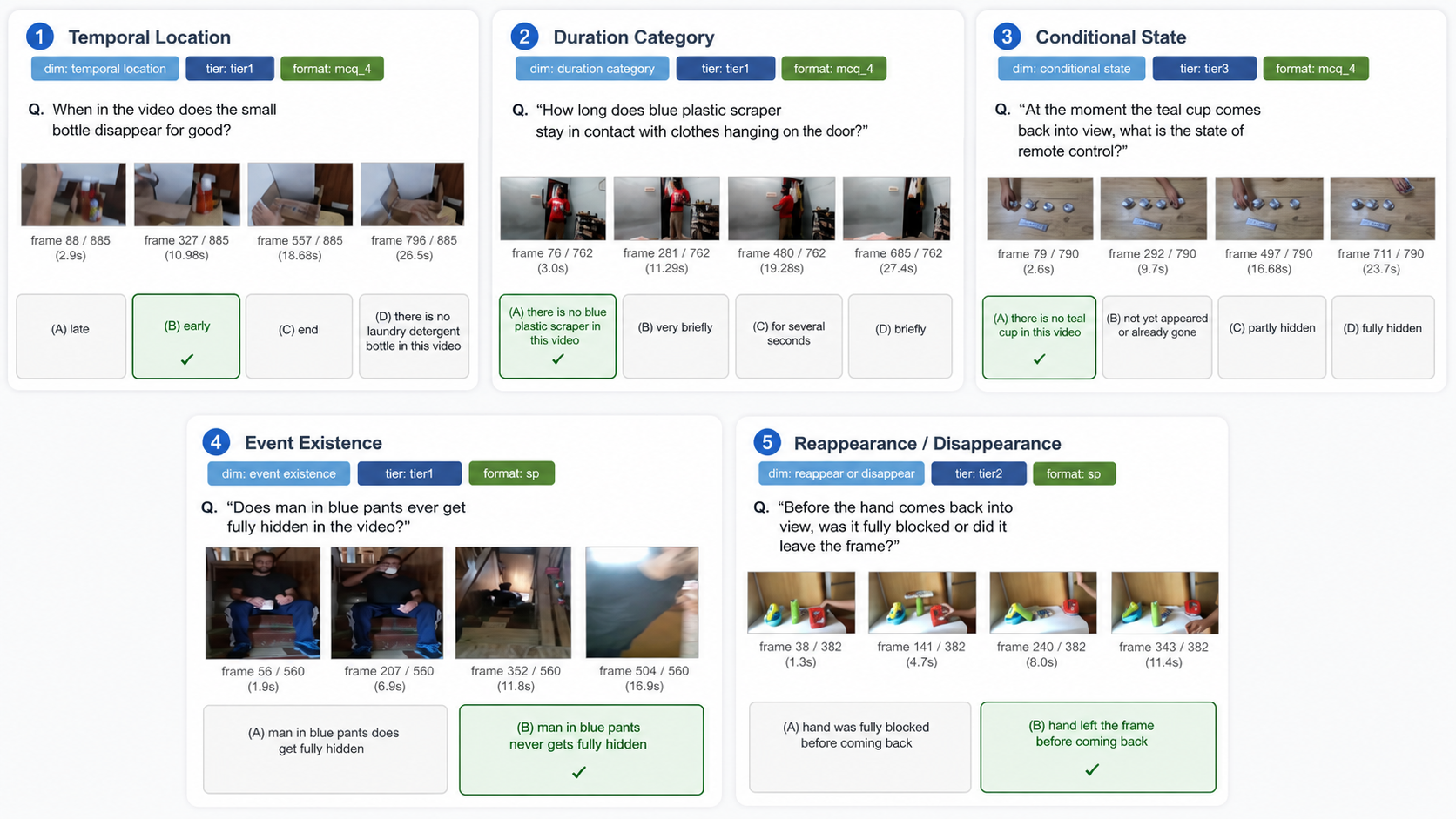}
    \caption{
    Dimension-wise qualitative examples from TOC-Bench, Part I.
    The examples cover temporal location, duration category, conditional state, event existence, and reappearance/disappearance.
    They illustrate how TOC-Bench tests object-level temporal reasoning over disappearance timing, duration estimation, conditional object state, event existence, and occlusion-versus-leaving-frame distinctions.
    }
    \label{fig:dimension-examples-part1}
\end{figure*}

\begin{figure*}[t]
    \centering
    \includegraphics[width=\textwidth]{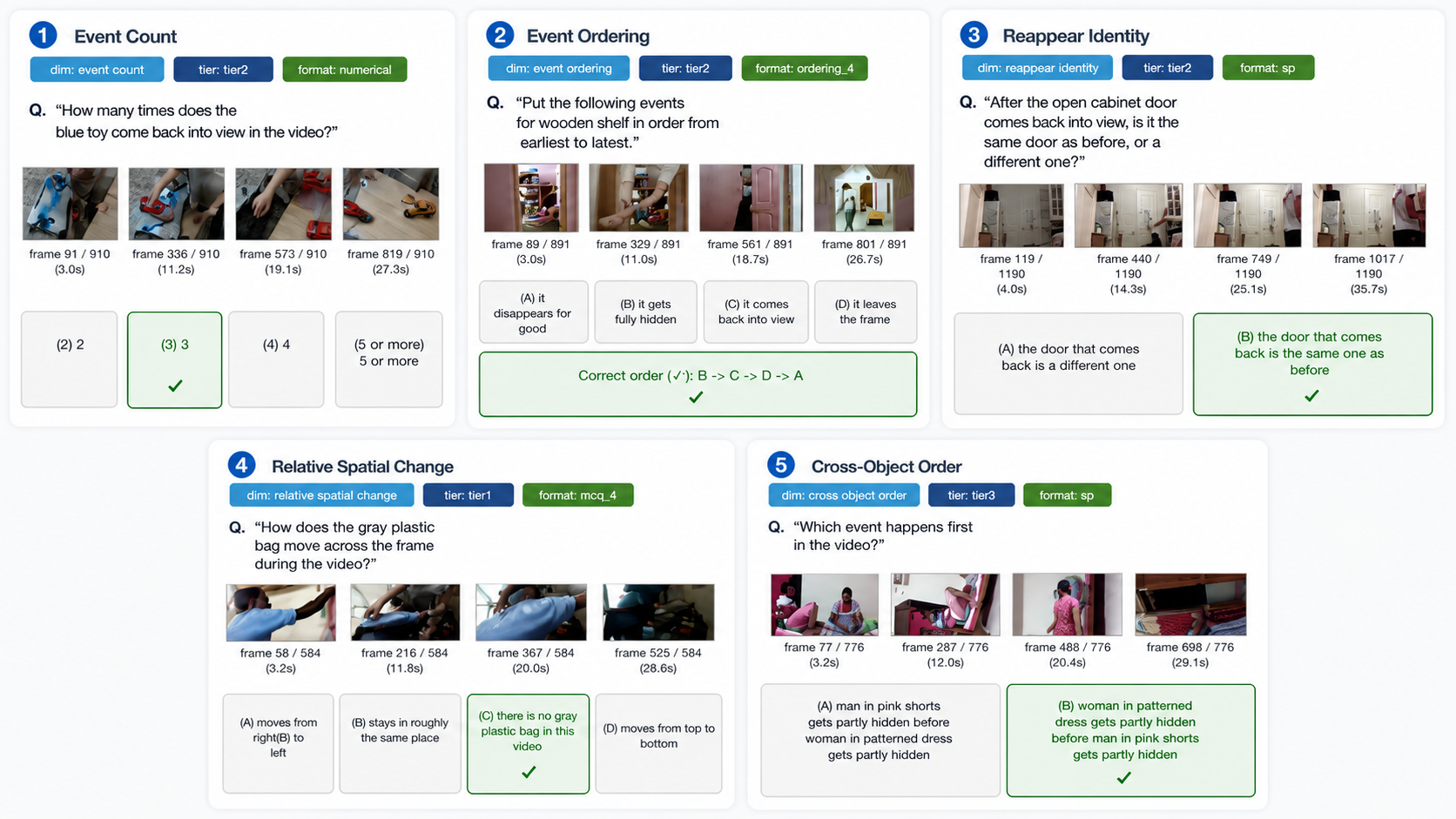}
    \caption{
    Dimension-wise qualitative examples from TOC-Bench, Part II.
    The examples cover cross-object order, event count, and event ordering.
    They illustrate diagnostic questions that require comparing events across different objects, accumulating repeated object-level events, and reconstructing chronological event sequences from temporally ordered visual evidence.
    }
    \label{fig:dimension-examples-part2}
\end{figure*}

Together, \cref{fig:dimension-examples-part1,fig:dimension-examples-part2} show that the dimensions in TOC-Bench are not merely different surface templates. They correspond to distinct temporal object-consistency skills, including locating when an object event occurs, estimating event duration, verifying event existence, distinguishing occlusion from leaving the frame, counting repeated events, and reasoning over event order within or across object tracks.

\subsection{Answer-Balance Analysis}

\begin{figure}[t]
    \centering
    \includegraphics[width=0.65\linewidth]{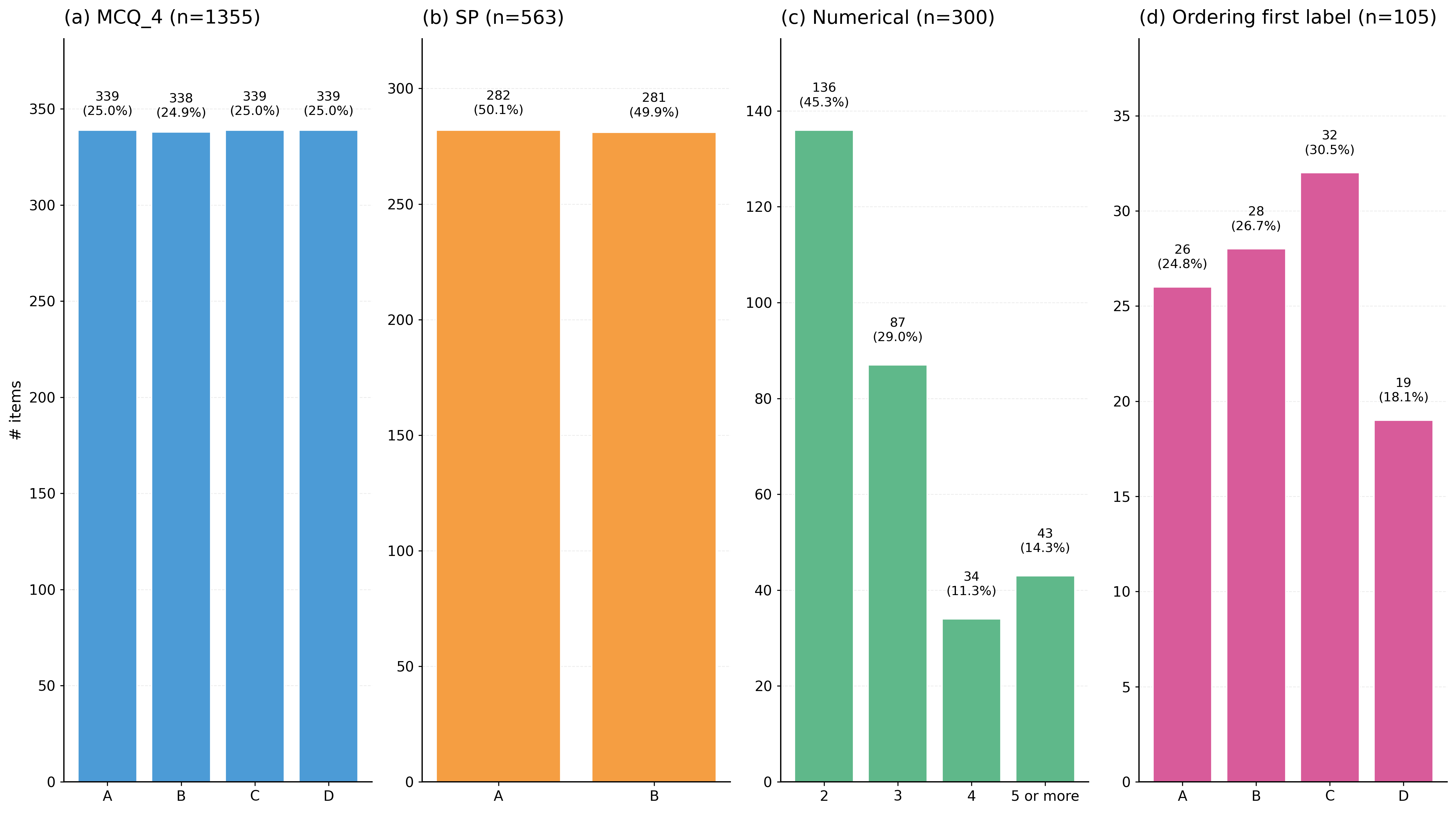}
    \caption{
    Answer-balance audit across question formats.
    The correct labels are approximately balanced for multiple-choice and statement-pair questions, while numerical and ordering questions do not admit a strong majority-class shortcut.
    }
    \label{fig:answer-balance}
\end{figure}

\Cref{fig:answer-balance} shows whether TOC-Bench contains trivial majority-label shortcuts. The four-way multiple-choice format is nearly uniformly balanced across answer labels, and the statement-pair format is close to a 50/50 split. Numerical and ordering questions also avoid a single dominant answer class. This reduces the risk that a model can obtain high accuracy by exploiting answer-label priors alone.

\subsection{Subject-Reference Diversity}

\begin{figure}[t]
    \centering
    \includegraphics[width=0.7\linewidth]{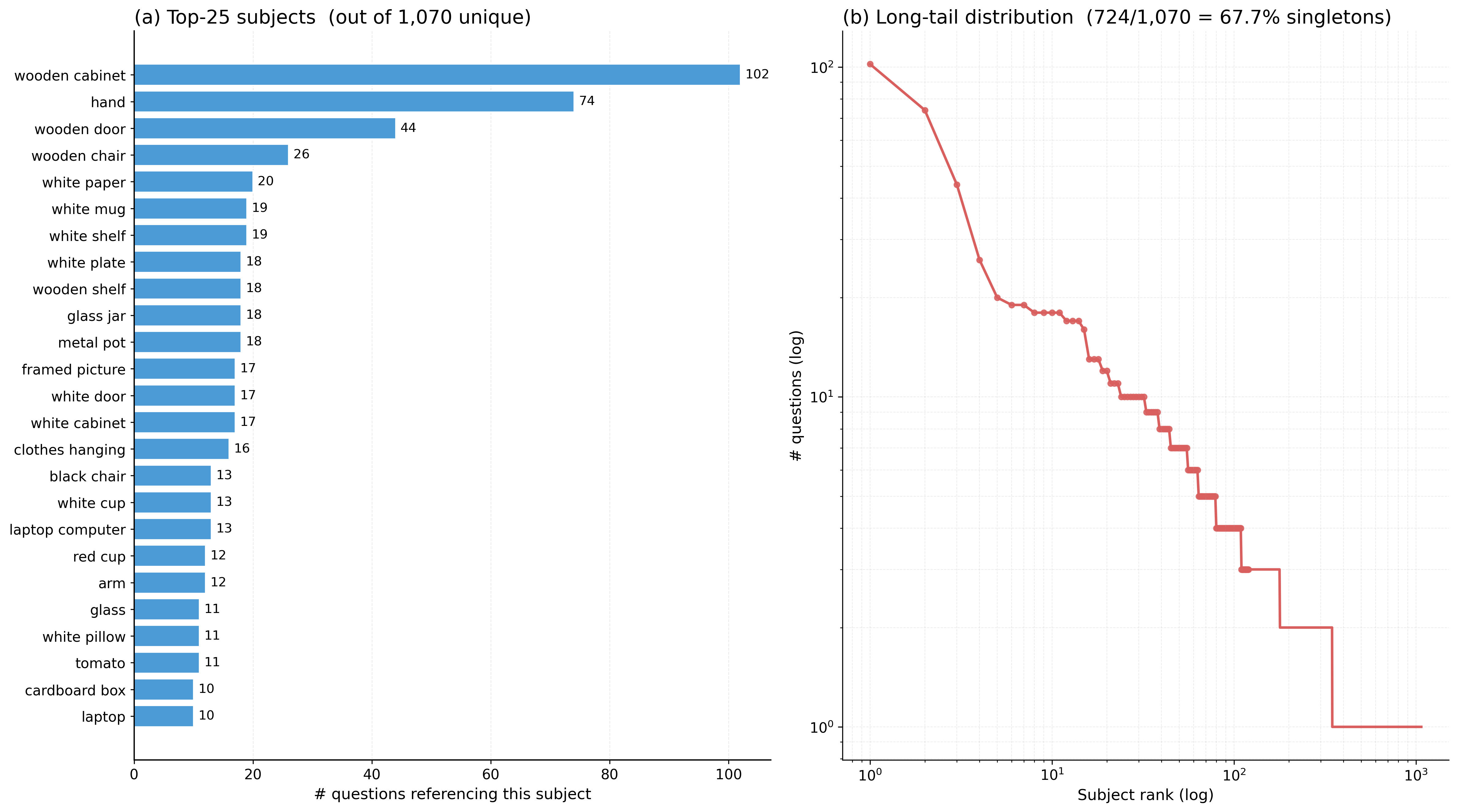}
    \caption{
    Subject-reference diversity in TOC-Bench.
    The left panel shows the most frequent subject references, while the right panel shows a long-tailed distribution over 1,071 unique subject references.
    }
    \label{fig:subject-diversity}
\end{figure}

\Cref{fig:subject-diversity} shows that TOC-Bench spans a long tail of subject references. Although common objects such as hands, cabinets, doors, and chairs appear frequently, the benchmark contains 1,071 unique subject references, with most appearing only once. This long-tailed distribution helps prevent the benchmark from collapsing into a small set of frequent object categories.

\subsection{Length-Leakage Diagnostics}

\begin{figure}[t]
    \centering
    \includegraphics[width=0.65\linewidth]{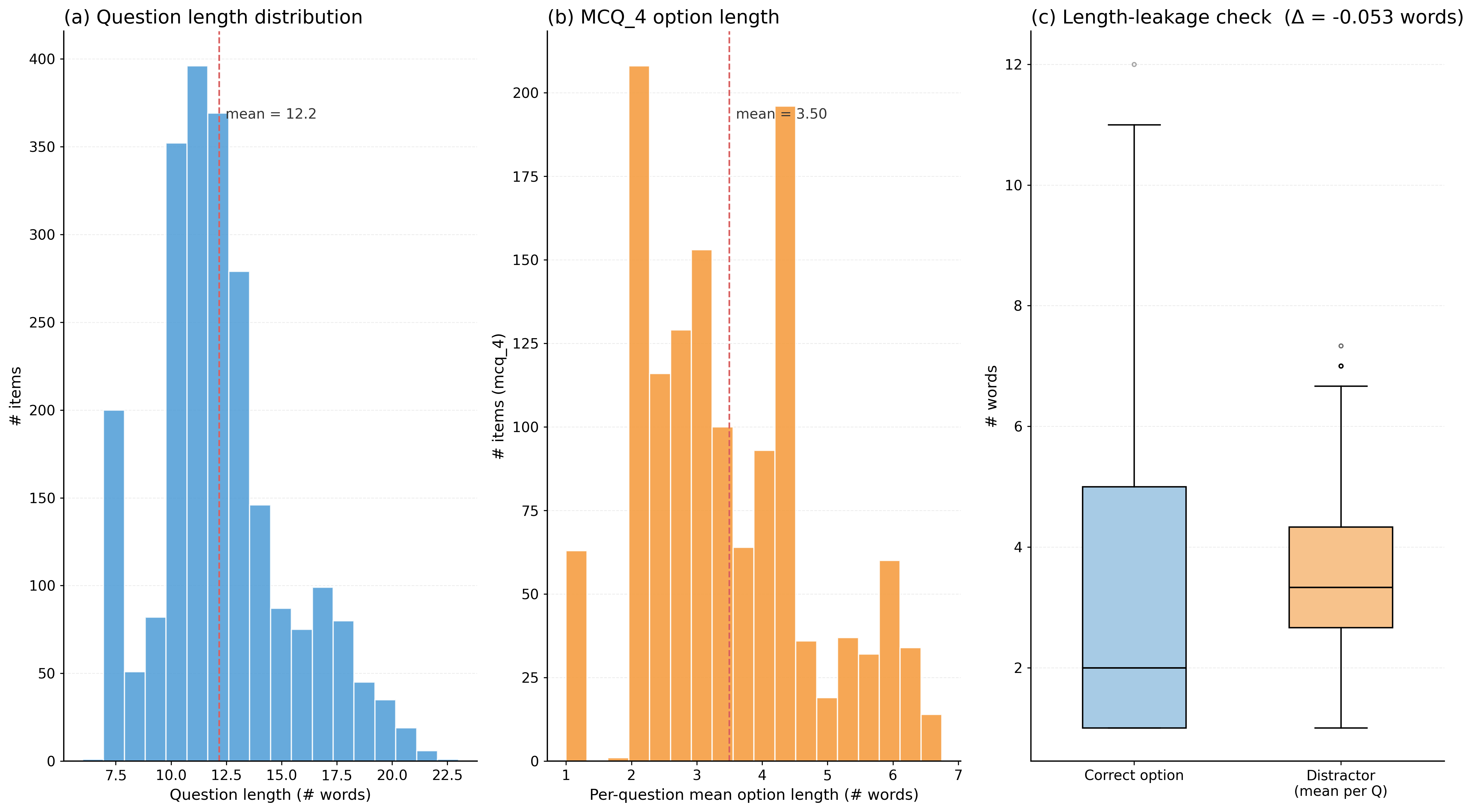}
    \caption{
    Length-leakage diagnostics for TOC-Bench.
    The analysis compares question lengths, multiple-choice option lengths, and the length difference between correct options and distractors.
    The correct option is not systematically longer or shorter than the distractors.
    }
    \label{fig:length-leakage}
\end{figure}

\Cref{fig:length-leakage} checks whether the dataset contains obvious length-based shortcuts. The average question length is moderate, and multiple-choice options have compact and comparable lengths. More importantly, the correct option is not systematically longer or shorter than the distractor options, suggesting that option length is unlikely to provide a strong shortcut for answering TOC-Bench questions.
\section{Evaluation Protocol and Frame Sampling}
\label{appendix:frame-protocol}

TOC-Bench uses a controlled frame-based evaluation protocol. As reported in \cref{tab:tocbench-main-results}, most evaluated models receive 32 uniformly sampled frames. A small number of model-specific or legacy baselines use different feasible settings: LLaVA-Video-7B and Video-LLaVA-7B use 8 frames, LLaVA-OV-7B-1Frame uses a single frame, and Video-ChatGPT-7B uses 100 frames following its evaluation convention. We report the frame budget for each model explicitly so that the results are interpreted under the corresponding input setting.

We use extracted frames rather than direct video-file inputs as the main protocol for comparability and reproducibility. Different model families support different video interfaces: some closed-source APIs accept video files and perform hidden decoding, frame selection, temporal compression, and visual-token allocation, while many open-source Video-LLMs require pre-extracted frames and explicit visual-token budgets. If each model were evaluated only through its preferred input interface, the comparison would mix model capability with API-specific preprocessing and inaccessible internal sampling strategies. A frame-based protocol makes the visual evidence budget explicit and easier to reproduce.

We do not claim that a fixed number of sampled frames is the optimal input setting for every model. Rather, TOC-Bench evaluates temporal object consistency under a clearly specified visual input protocol. This is a common and practical setting for Video-LLMs, since long videos are often processed through sparse frame sampling or visual-token compression. The results should therefore be interpreted as measuring how well models preserve object identity, state, and event relations under the reported frame budgets, rather than as an exhaustive estimate of each model's best possible performance under all available input modes.

We use uniform sampling instead of event-aware sampling to avoid leaking benchmark construction information into the model input. Since TOC-Bench is built from object tracks and event timelines, selecting frames around ground-truth events would make the input depend on the answer. Uniform sampling is model-agnostic and does not use object tracks, event labels, or question-specific temporal metadata at evaluation time.
\section{Additional Evaluation Results}
\label{appendix:additional-results}

\subsection{Tier-Level and Format-Level Results}
\label{appendix:tier-format-results}

\begin{table*}[t]
\caption{
Tier-level and format-level results on TOC-Bench.
Tier 1 focuses on single-object temporal reasoning, Tier 2 requires cross-time identity consistency for one object, and Tier 3 requires cross-object temporal reasoning.
MCQ denotes four-way multiple-choice questions, SP denotes statement-pair questions, Num denotes numerical event-count questions, and Ord-3/Ord-4 denote ordering questions with three or four events.
}
\centering
\scriptsize
\setlength{\tabcolsep}{2pt}
\renewcommand{\arraystretch}{0.55}
\resizebox{1.0\textwidth}{!}{
\begin{tabular}{lcccccccc}
\toprule
Model & Tier 1 & Tier 2 & Tier 3 & MCQ & SP & Num & Ord-3 & Ord-4 \\
\midrule
Human & 92.4 & 81.9 & 89.7 & 91.9 & 87.7 & 84.0 & 80.8 & 79.2 \\
\midrule
\multicolumn{9}{l}{\textit{Closed-source models}} \\
GPT-5.5 & 50.4 & 36.8 & 50.3 & 45.6 & 70.2 & 18.3 & 19.2 & 30.2 \\
Kimi-K2.6 & 44.3 & 34.7 & 59.4 & 44.8 & 63.2 & 19.3 & 23.1 & 22.6 \\
Gemini-3.1-Pro-Preview & 45.5 & 32.7 & 52.9 & 44.7 & 60.4 & 18.7 & 21.2 & 13.2 \\
Grok-4.3 & 43.7 & 34.5 & 55.5 & 44.5 & 60.0 & 19.7 & 19.2 & 15.1 \\
Seed-2.0-Lite & 44.1 & 32.7 & 47.3 & 41.8 & 55.6 & 24.3 & 23.1 & 20.8 \\
GLM-5V-Turbo & 39.3 & 29.2 & 44.5 & 36.7 & 57.5 & 13.7 & 19.2 & 15.1 \\
Gemini-3.1-Flash-Lite-Preview & 34.8 & 23.8 & 43.0 & 35.3 & 48.3 & 6.0 & 19.2 & 13.2 \\
GPT-5.4-mini & 32.4 & 27.8 & 41.5 & 31.0 & 48.5 & 21.3 & 17.3 & 5.7 \\
Mimo-V2-Omni & 31.9 & 20.5 & 44.3 & 31.5 & 50.8 & 4.7 & 13.5 & 1.9 \\
\midrule
\multicolumn{9}{l}{\textit{Open-source thinking / reasoning models}} \\
Qwen3-VL-8B-Thinking & 31.9 & 21.9 & 44.7 & 33.1 & 43.2 & 14.0 & 15.4 & 5.7 \\
VideoChat-R1.5-7B & 28.2 & 20.8 & 43.2 & 28.0 & 48.5 & 7.0 & 19.2 & 0.0 \\
Video-R1-7B & 24.4 & 14.6 & 39.8 & 25.0 & 40.7 & 0.0 & 23.1 & 5.7 \\
\midrule
\multicolumn{9}{l}{\textit{Open-source standard models}} \\
Qwen2.5-VL-72B & 36.3 & 20.1 & 44.5 & 35.4 & 50.8 & 3.0 & 19.2 & 11.3 \\
InternVL3-8B & 30.2 & 24.9 & 36.3 & 28.5 & 44.8 & 18.7 & 9.6 & 1.9 \\
LLaVA-Video-72B & 28.5 & 23.0 & 37.2 & 25.8 & 46.5 & 15.0 & 21.2 & 7.5 \\
VideoLLaMA3-7B & 25.8 & 31.5 & 31.8 & 23.5 & 39.8 & 33.7 & 25.0 & 7.5 \\
LLaVA-Video-7B & 27.2 & 27.4 & 30.5 & 24.6 & 40.3 & 26.3 & 9.6 & 9.4 \\
Video-LLaVA-7B & 28.9 & 17.8 & 35.9 & 25.2 & 50.1 & 4.7 & 5.8 & 1.9 \\
MiniCPM-V-2.6 & 24.2 & 26.3 & 36.1 & 22.5 & 45.8 & 18.7 & 13.5 & 5.7 \\
Qwen2.5-VL-7B & 25.8 & 16.2 & 43.7 & 27.1 & 43.3 & 0.7 & 25.0 & 3.8 \\
LLaVA-OV-7B-1Frame & 23.4 & 29.2 & 26.7 & 21.5 & 36.2 & 28.3 & 11.5 & 7.5 \\
Qwen3-VL-8B & 22.1 & 22.6 & 36.8 & 21.8 & 42.6 & 12.3 & 15.4 & 7.5 \\
Video-ChatGPT-7B & 20.4 & 29.2 & 12.7 & 18.4 & 24.7 & 30.7 & 13.5 & 1.9 \\
\bottomrule
\end{tabular}
}
\label{tab:appendix-tier-format-results}
\end{table*}

\Cref{tab:appendix-tier-format-results} provides a more detailed breakdown by difficulty tier and task format. The results show that Tier 2 is consistently difficult for many models, especially because it contains identity-sensitive and repeated-event questions for the same object. The format-level results also show that statement-pair questions are generally easier than numerical counting and event ordering. This supports the main-text observation that current Video-LLMs are better at binary temporal judgments than at structured temporal accumulation and chronological reconstruction.

\subsection{Hallucination-Bucket Results}
\label{appendix:hallucination-results}

\begin{table*}[t]
\caption{
Accuracy on hallucination-related buckets.
Clean denotes questions without hallucination injection.
Distractor-only denotes questions with a hallucination-style wrong option.
Variant A refers to nonexistent-subject questions, and Variant B refers to existing-subject but nonexistent-event questions.
}
\centering
\tiny
\setlength{\tabcolsep}{2pt}
\renewcommand{\arraystretch}{0.45}
\resizebox{1.0\textwidth}{!}{
\begin{tabular}{lcccc}
\toprule
Model & Clean & Distractor-only & Variant A & Variant B \\
\midrule
Human & 89.3 & 89.3 & 89.4 & 89.5 \\
\midrule
\multicolumn{5}{l}{\textit{Closed-source models}} \\
GPT-5.5 & 48.2 & 48.3 & 31.7 & 50.9 \\
Kimi-K2.6 & 44.9 & 39.5 & 60.6 & 47.4 \\
Gemini-3.1-Pro-Preview & 42.2 & 44.1 & 50.0 & 46.0 \\
Grok-4.3 & 42.5 & 39.0 & 59.6 & 49.5 \\
Seed-2.0-Lite & 41.9 & 42.7 & 32.7 & 47.4 \\
GLM-5V-Turbo & 38.8 & 38.7 & 26.9 & 40.4 \\
Gemini-3.1-Flash-Lite-Preview & 31.9 & 31.6 & 42.8 & 40.0 \\
GPT-5.4-mini & 34.9 & 33.5 & 18.3 & 35.8 \\
Mimo-V2-Omni & 31.5 & 29.5 & 40.9 & 30.9 \\
\midrule
\multicolumn{5}{l}{\textit{Open-source thinking / reasoning models}} \\
Qwen3-VL-8B-Thinking & 30.5 & 26.4 & 48.6 & 40.4 \\
VideoChat-R1.5-7B & 30.5 & 25.3 & 30.8 & 34.7 \\
Video-R1-7B & 25.1 & 19.9 & 27.9 & 35.8 \\
\midrule
\multicolumn{5}{l}{\textit{Open-source standard models}} \\
Qwen2.5-VL-72B & 31.5 & 32.1 & 40.9 & 43.9 \\
InternVL3-8B & 31.6 & 29.5 & 20.7 & 33.0 \\
LLaVA-Video-72B & 31.8 & 27.4 & 18.8 & 28.8 \\
VideoLLaMA3-7B & 32.9 & 23.9 & 17.8 & 29.8 \\
LLaVA-Video-7B & 30.8 & 27.7 & 15.9 & 26.0 \\
Video-LLaVA-7B & 30.2 & 24.9 & 26.0 & 25.3 \\
MiniCPM-V-2.6 & 31.9 & 22.7 & 14.9 & 27.7 \\
Qwen2.5-VL-7B & 26.8 & 21.6 & 33.7 & 36.5 \\
LLaVA-OV-7B-1Frame & 28.3 & 19.2 & 33.2 & 23.9 \\
Qwen3-VL-8B & 29.3 & 18.5 & 21.2 & 28.4 \\
Video-ChatGPT-7B & 26.5 & 18.4 & 15.9 & 9.8 \\
\bottomrule
\end{tabular}
}
\label{tab:appendix-hallucination-results}
\end{table*}

\Cref{tab:appendix-hallucination-results} shows that different models have different hallucination robustness profiles. For example, GPT-5.5 performs strongly on clean and distractor-only items, but drops on nonexistent-subject questions. In contrast, Kimi-K2.6 and Grok-4.3 perform much better on Variant A, suggesting stronger ability to reject nonexistent subjects. Seed-2.0-Lite shows a different pattern, performing better on nonexistent-event questions than on nonexistent-subject questions. These differences explain why HDA is not always aligned with overall accuracy and justify treating hallucination robustness as a separate diagnostic axis in TOC-Bench.

\subsection{Additional Visualizations}
\label{appendix:result-visualizations}

\begin{figure*}[t]
    \centering
    \includegraphics[width=0.85\textwidth]{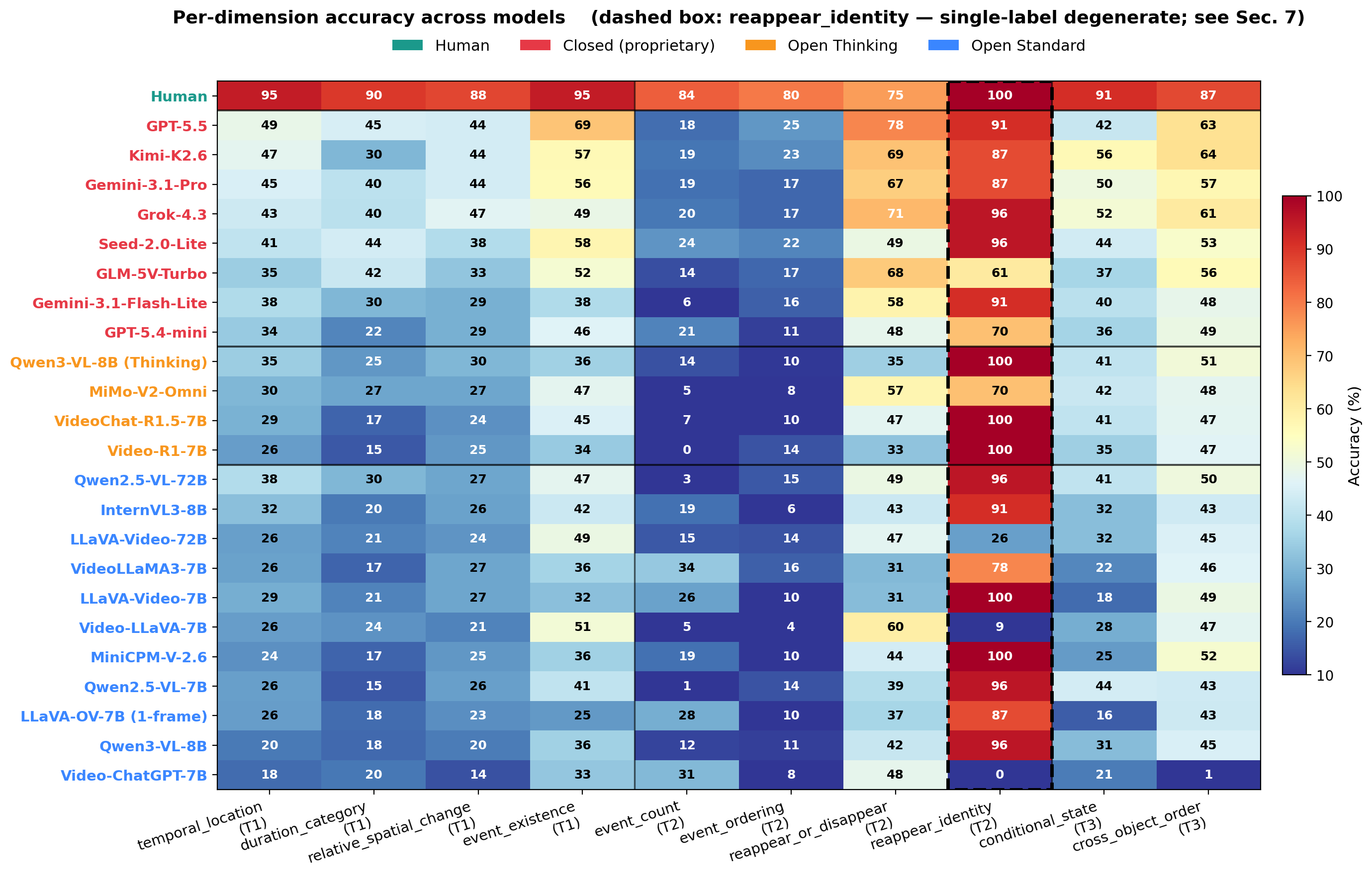}
    \caption{
    Model-by-dimension accuracy heatmap.
    The heatmap highlights that event counting and event ordering are consistently difficult across model families, while statement-style dimensions such as event existence and cross-object order are relatively easier.
    }
    \label{fig:model-dimension-heatmap}
\end{figure*}

\paragraph{Model-by-dimension behavior.}
\Cref{fig:model-dimension-heatmap} provides a dense view of per-dimension accuracy across all evaluated models. The heatmap makes two patterns clear. First, human performance is consistently high across all 10 dimensions, while all model families show large drops on several object-level temporal reasoning dimensions. Second, event counting and event ordering are difficult across almost all models, suggesting that current Video-LLMs struggle with structured temporal accumulation and chronological reconstruction even when they perform better on event existence or statement-style dimensions.

\begin{figure*}[t]
    \centering
    \includegraphics[width=0.85\textwidth]{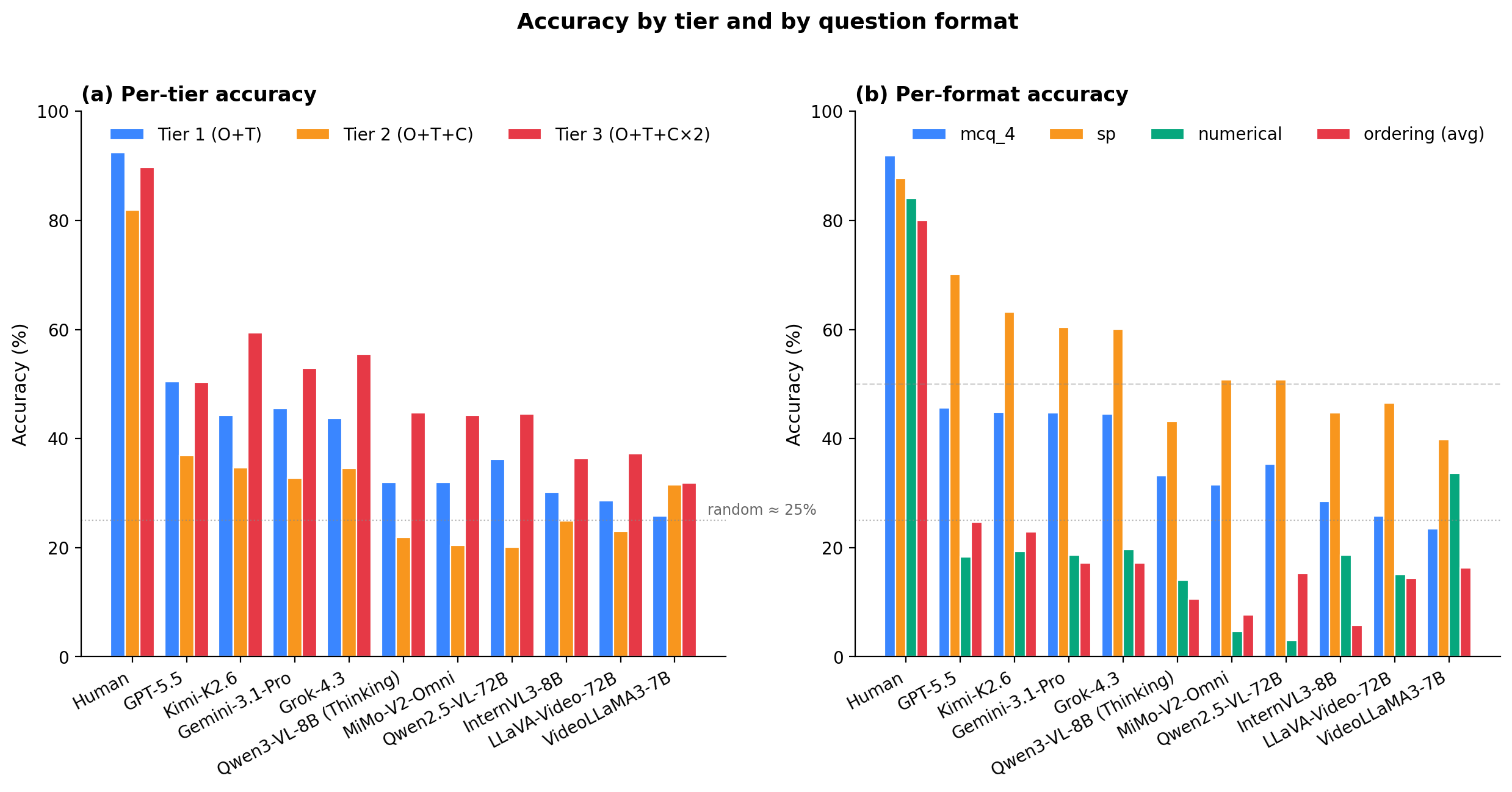}
    \caption{
    Tier-level and format-level accuracy breakdowns.
    Models generally perform better on statement-pair questions than on numerical counting and ordering formats.
    }
    \label{fig:tier-format-results}
\end{figure*}

\paragraph{Tier and format effects.}
\Cref{fig:tier-format-results} summarizes model performance by difficulty tier and task format. The tier-level view shows that Tier 2 remains challenging for many models because it requires cross-time consistency for the same object. The format-level view shows that statement-pair questions are generally easier than numerical counting and ordering questions. This supports the main finding that current models can often make binary temporal judgments, but are less reliable when they need to count repeated object events or recover an exact event order.

\begin{figure*}[t]
    \centering
    \includegraphics[width=0.85\textwidth]{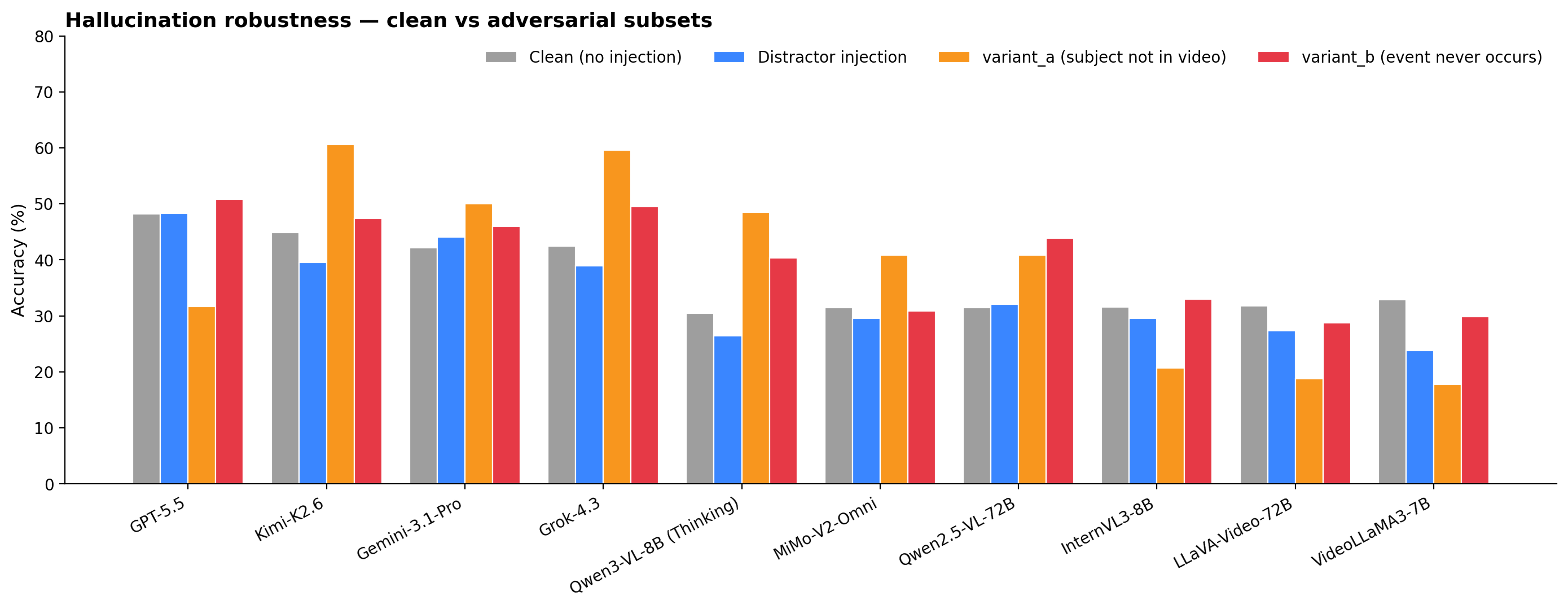}
    \caption{
    Hallucination-bucket accuracy across models.
    Different models show different robustness patterns on clean, distractor-only, nonexistent-subject, and nonexistent-event questions.
    }
    \label{fig:hallucination-results}
\end{figure*}

\paragraph{Hallucination-aware behavior.}
\Cref{fig:hallucination-results} compares model performance across clean questions, hallucination-style distractor questions, nonexistent-subject variants, and nonexistent-event variants. The results show that hallucination robustness varies substantially across models. Some models perform relatively well on clean or distractor-only items but drop on nonexistent-subject variants, while others are better at rejecting nonexistent subjects but less stable on nonexistent events. This confirms that hallucination-aware construction provides information that is not visible from overall accuracy alone.

\begin{figure*}[t]
    \centering
    \includegraphics[width=0.85\textwidth]{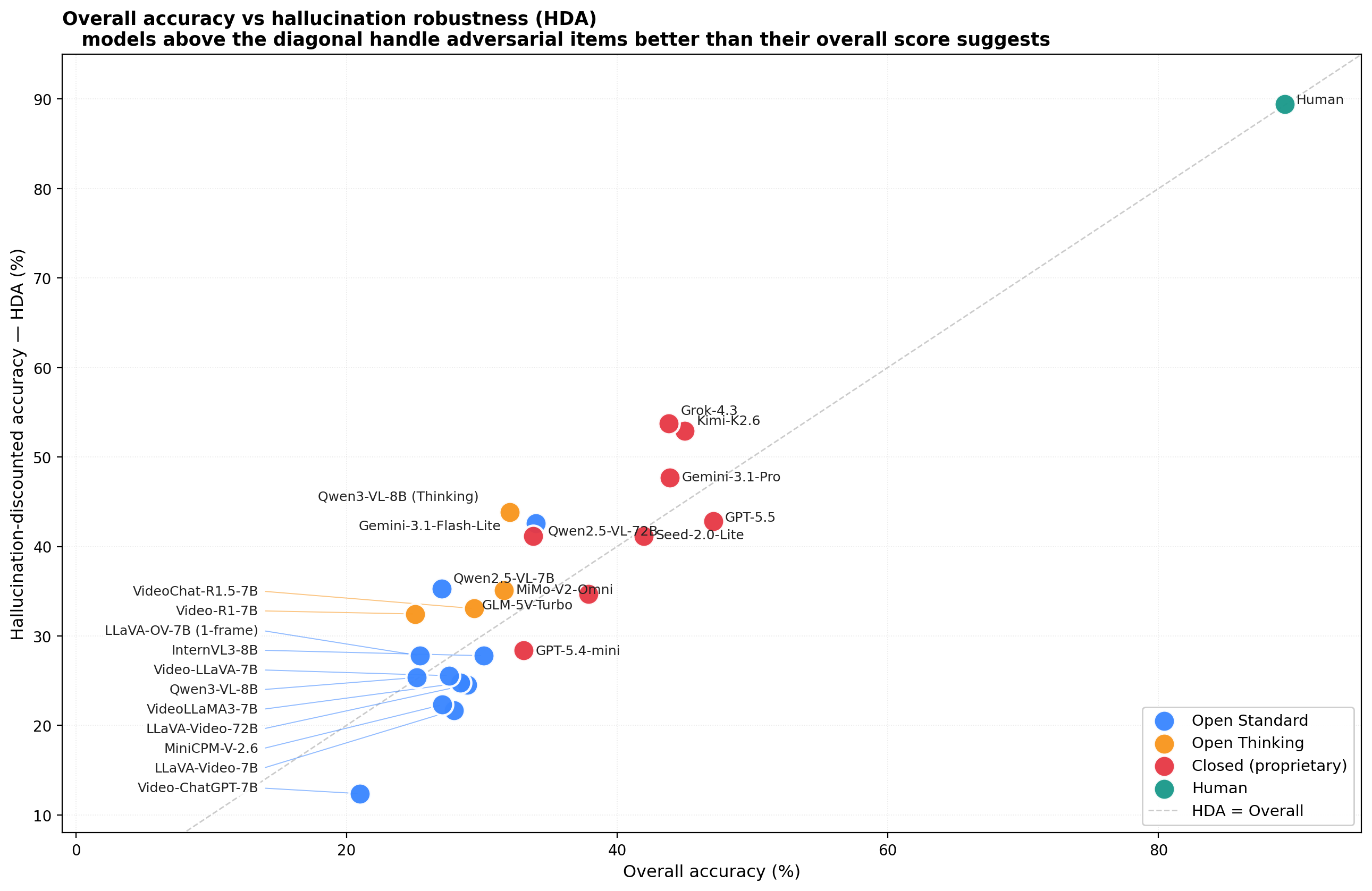}
    \caption{
    Overall accuracy versus hallucination diagnostic accuracy.
    HDA is not perfectly aligned with overall accuracy, suggesting that object/event hallucination robustness is a distinct evaluation axis in TOC-Bench.
    }
    \label{fig:overall-hda-scatter}
\end{figure*}

\paragraph{Overall accuracy versus HDA.}
Finally, \cref{fig:overall-hda-scatter} directly compares overall accuracy with hallucination diagnostic accuracy. The two metrics are positively related but not perfectly aligned. For example, some models with strong overall accuracy do not achieve the highest HDA, while models such as Grok-4.3 and Kimi-K2.6 show stronger hallucination-aware performance than their overall rank alone would suggest. This supports our use of HDA as a separate diagnostic metric for object/event hallucination robustness.

\subsection{Qualitative Failure Cases}
\label{appendix:failure-cases}

In addition to quantitative breakdowns, we provide qualitative failure cases to illustrate common error modes exposed by TOC-Bench. These examples show that models often fail not because the target object is absent from the video, but because the question requires tracking object events over time, distinguishing similar temporal states, or rejecting plausible but unsupported assumptions.

\begin{figure*}[t]
    \centering
    \includegraphics[width=\textwidth]{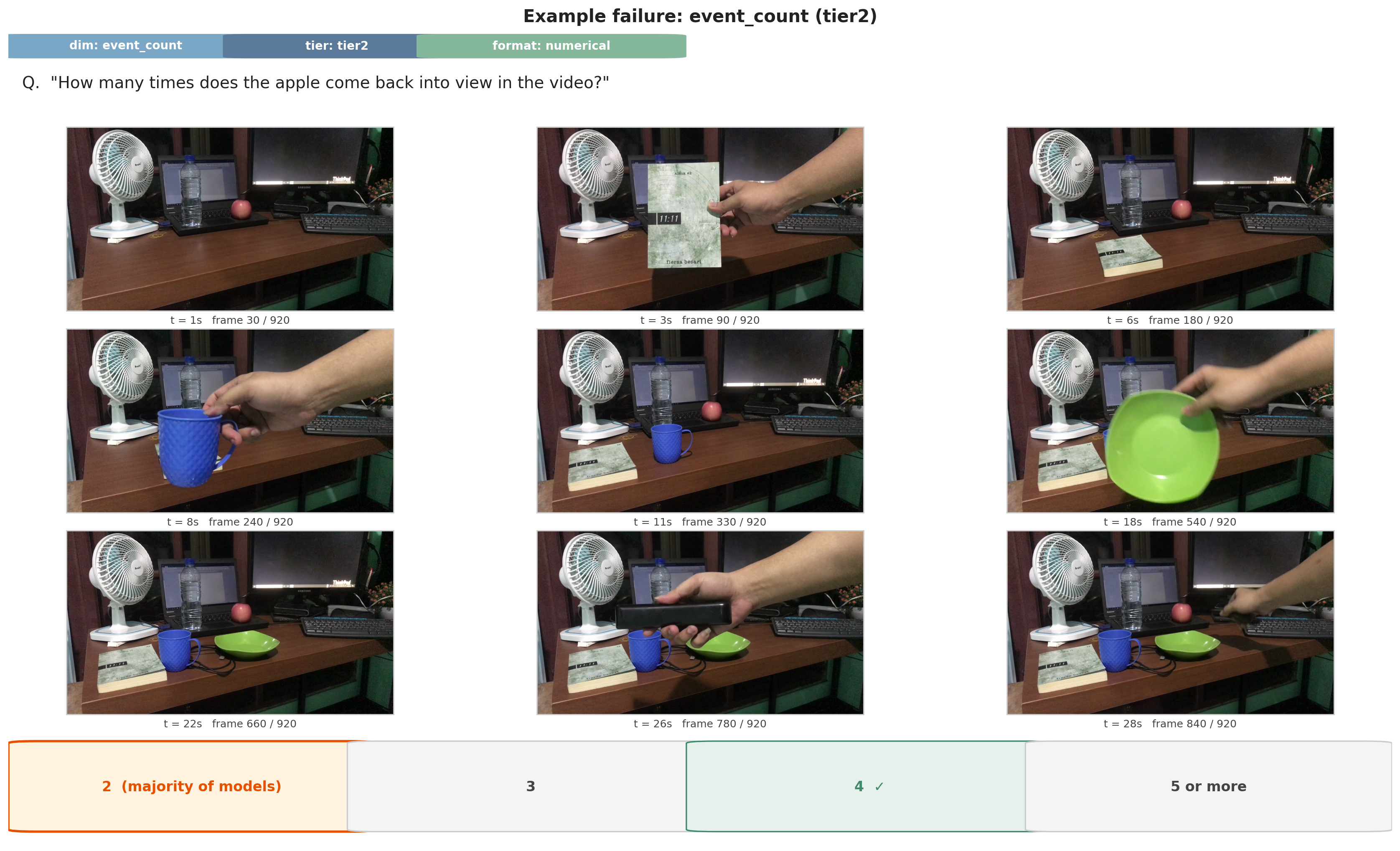}
    \caption{
    Failure case on \textbf{event counting}.
    The question asks how many times the apple comes back into view. 
    The correct answer is 4, but the majority of evaluated models predict 2.
    This example shows that models can recognize the target object and its reappearance events locally, yet still fail to accumulate repeated object-level events across the full video.
    }
    \label{fig:failure-event-count}
\end{figure*}

\begin{figure*}[t]
    \centering
    \includegraphics[width=\textwidth]{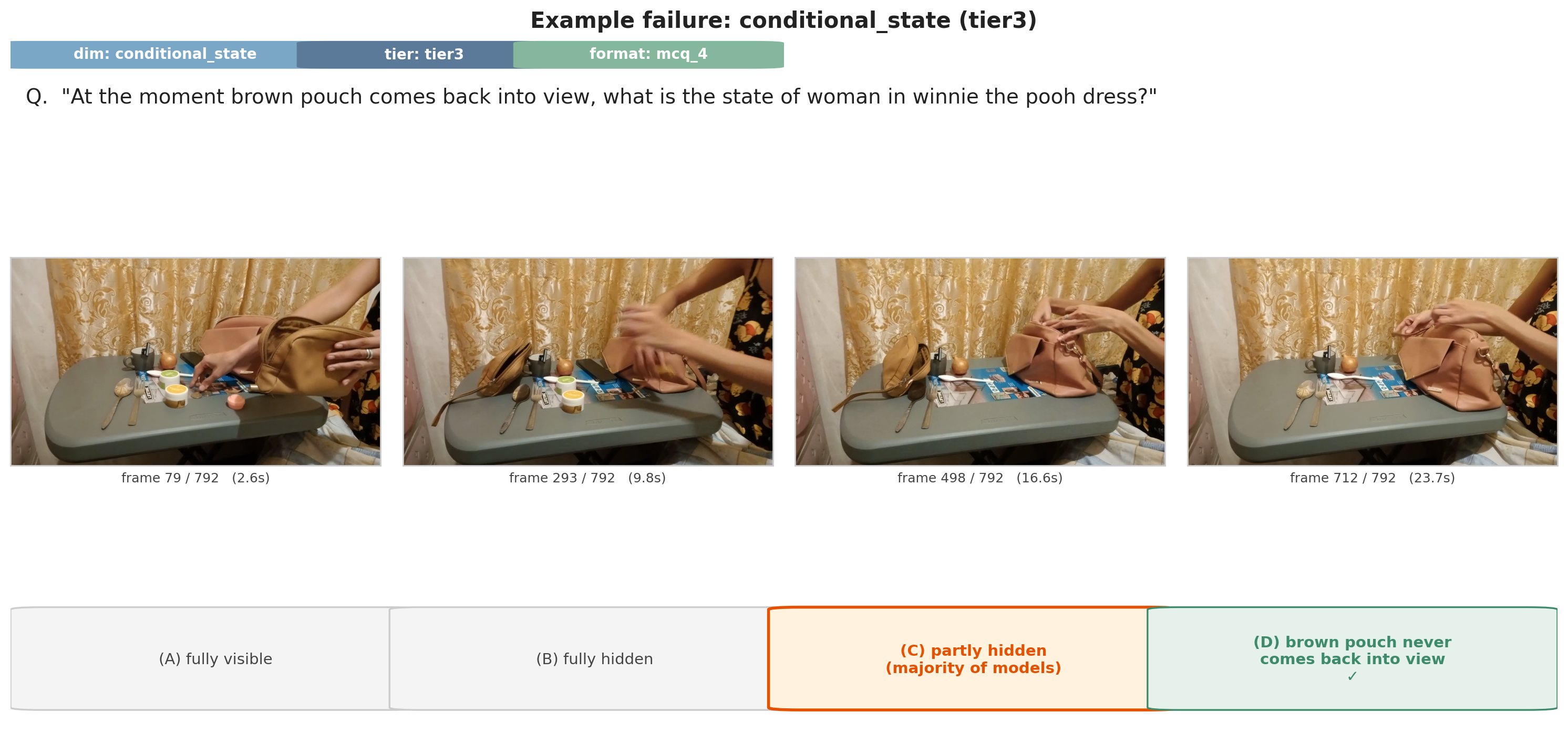}
    \caption{
    Failure case on \textbf{conditional state reasoning}.
    The question asks for the state of the woman at the moment when the brown pouch comes back into view.
    The correct answer is that the brown pouch never comes back into view, while the majority of models choose a plausible visual state of the woman.
    This example illustrates a hallucination-aware temporal failure: models tend to assume the queried event happens and answer the conditional part, instead of rejecting the false temporal premise.
    }
    \label{fig:failure-conditional-state}
\end{figure*}

As shown in \cref{fig:failure-event-count,fig:failure-conditional-state}, the two cases reveal different failure modes. The first case corresponds to a repeated-event accumulation failure: although the apple is visible in multiple sampled frames, correctly answering the question requires counting all reappearance events rather than relying on a small number of salient moments. The second case corresponds to a conditional hallucination failure: the model must first verify whether the triggering event actually occurs; otherwise, selecting the state of the other object gives a plausible but incorrect answer. Together, these cases complement the quantitative results in \cref{tab:tocbench-main-results}, showing why event counting and hallucination-aware verification remain difficult for current Video-LLMs.

\clearpage


\end{document}